\documentclass{article}

\usepackage{arxiv}

\usepackage[utf8]{inputenc} 
\usepackage[T1]{fontenc}    
\usepackage{url}            
\usepackage{amsfonts}       
\usepackage{nicefrac}       
\usepackage{microtype}      
\usepackage{graphicx}
\usepackage{bm}
\usepackage{amsmath}
\usepackage{algorithm}
\usepackage{algpseudocode}
\usepackage{subcaption}
\usepackage{booktabs}
\usepackage{multirow}
\usepackage{hyperref} 
\usepackage[numbers]{natbib}
\usepackage{pifont}
\usepackage{xcolor}
\usepackage{float}

\title{Zero-Shot Solving of Imaging Inverse Problems via Noise-Refined Likelihood Guided Diffusion Models}

\author{
  Zhen Wang \\
  School of Mathematics and Statistics \\
  Nanjing University of Science and Technology \\
  \texttt{zhenw@njust.edu.cn}
  \and
  Hongyi Liu \thanks{Corresponding author: hyliu@njust.edu.cn}\\
  School of Mathematics and Statistics \\
  Nanjing University of Science and Technology \\
  \texttt{hyliu@njust.edu.cn}
  \and
  Zhihui Wei \\
  School of Computer Science and Engineering \\
  Nanjing University of Science and Technology \\
  \texttt{gswei@njust.edu.cn}
}

\renewcommand{\shorttitle}{Diffusion Models with Noise-Refined Likelihood Guidance for Imaging Inverse Problems}

\begin{document}
\maketitle

\begin{abstract}
	Diffusion models have achieved remarkable success in imaging inverse problems owing to their powerful generative capabilities. However, existing approaches typically rely on models trained for specific degradation types, limiting their generalizability to various degradation scenarios. To address this limitation, we propose a zero-shot framework capable of handling various imaging inverse problems without model retraining. We introduce a likelihood-guided noise refinement mechanism that derives a closed-form approximation of the likelihood score, simplifying score estimation and avoiding expensive gradient computations. This estimated score is subsequently utilized to refine the model-predicted noise, thereby better aligning the restoration process with the generative framework of diffusion models. In addition, we integrate the Denoising Diffusion Implicit Models (DDIM) sampling strategy to further improve inference efficiency. The proposed mechanism can be applied to both optimization-based and sampling-based schemes, providing an effective and flexible zero-shot solution for imaging inverse problems. Extensive experiments demonstrate that our method achieves superior performance across multiple inverse problems, particularly in compressive sensing, delivering high-quality reconstructions even at an extremely low sampling rate (5\%).
\end{abstract}


\section{Introduction}\label{sec1}

Images often suffer from various degradations during acquisition, transmission, or storage due to hardware constraints and environmental factors, resulting in noise \cite{noise1,noise2,noise3,noise4}, blurring \cite{bluring1,bluring2,bluring3,bluring4} and reduced resolution \cite{sr1,sr2,sr3,ilvr}. Such distortions directly impact the performance of downstream applications. For example, in the medical field \cite{medical,medical2,medical3}, image quality crucially determines the reliability of clinical diagnoses, while in remote sensing \cite{hsi,marghany2022remote}, higher-quality images contain essential information for precise land cover classification and environmental monitoring. Consequently, restoring high-quality images from degraded observations has become a central topic in computer vision.

Mathematically, the degradation process of an image can be modeled as:
\begin{align}
    \bm{y} = \bm{Ax}+ \bm{z},
    \label{eq1}
\end{align}
where $ \bm{y} \in \mathbb{R}^{M}$ denotes the degraded measurement, $\bm{x} \in \mathbb{R}^{N}$ represents the clean image, $\bm{A} \in \mathbb{R}^{M\times N}$ characterizes the known linear degradation operator (e.g., downsampling matrix, blur kernel) and 
$\bm{z}\sim\mathcal{N}(\mathbf{0},\sigma_{y}^{2}\bm{I)}$ corresponds to additive random Gaussian noise.

Image restoration (IR) aims to estimate the original image $\bm{x}$ from its low-quality measurement $\bm{y}$, as formally described by the degraded model Eq.(\ref{eq1}). Due to the ill-posed nature of the degradation operator $\bm{A}$, this inverse problem results in a non-unique solution and sensitivity to noise perturbations. To constrain the solution space, a variational framework with regularization that incorporates the prior of $\bm{x}$ is typically written as:
\begin{align}
    \min_{\bm{x}} L(\bm{x}) = \underbrace{\ell(\bm{x},\bm{y})}_\text{data fidelity term} + \underbrace{s(\bm{x})}_{\text{prior term}},
    \label{eq2}
\end{align}
where $\ell(\mathbf{x},\mathbf{y})$ represents the data fidelity term, and $s(\bm{x})$ is the handcrafted regularizer.

\begin{figure}[t]
    \centering
    \includegraphics[width=\textwidth]{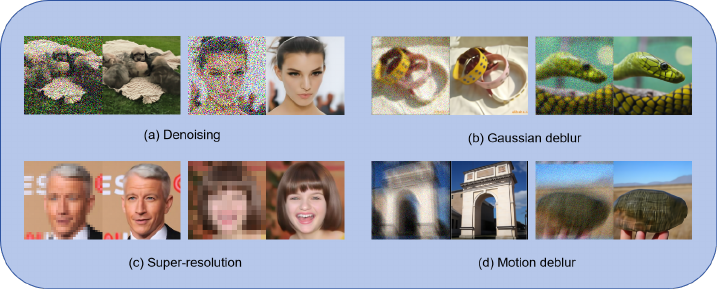} 
    \caption{Solving noisy linear problems with diffusion models. The restoration results (right) from the measurements (left) using our restoration method are shown.}
    \label{fig1} 
\end{figure}

Variational regularization methods are primarily based on physics-driven degradation modeling and handcrafted prior constraints (e.g., total variation (TV)\cite{rudin1992nonlinear}, sparsity\cite{sparse}, non-local\cite{shi2023geometric}), forming the foundation of classical frameworks. These methods typically minimize an energy functional to to trade off between data fidelity and prior constraints. For example, total variation (TV) regularization effectively preserves edges by penalizing excessive smoothness. Additionally, non-local techniques and sparse representation methods, such as the non-local means (NLM) algorithm \cite{shi2023geometric} and the BM3D algorithm \cite{noise3}, can be linked to variational principles, leveraging image self-similarity and sparse priors to enhance noise suppression and restoration performance.

However, these variational  methods (e.g., TV \cite{rudin1992nonlinear,cascarano2021combining,DING2022109751} and sparse priors \cite{sparse,lee2006efficient}) strongly depend on predefined priors. Compared to traditional spatial-domain methods (median filtering, bilateral filtering) and frequency-domain methods (wavelet thresholding \cite{wavelet}), they remain ineffective in recovering fine textures or capturing complex structural details. Moreover, these methods struggle to handle complex degradations (e.g., unknown noise mixtures, motion blur), and some of them are also computationally expensive, making real-time processing of large-scale images challenging.

With the development of deep learning, data-driven methods \cite{noise4,sr2,dl1,dl2} have achieved remarkable progress in the field of image restoration. Compared to traditional methods that rely on manually designed priors, deep learning methods learn the mapping between low-quality (LQ) and high-quality (HQ) images in an end-to-end manner, alleviating the dependence on explicit priors and allowing a more effective capture of complex degradation characteristics. Furthermore, the introduction of deep models such as convolutional neural networks (CNNs) and generative adversarial networks (GANs) has led to improvements in texture reconstruction, high-order structural preservation, and complex noise removal. However, despite these advances, deep learning-based methods remain challenged in effectively recovering fine details and higher-order structures, often producing overly smooth results. Additionally, these methods typically perform well only on specific degradations but exhibit poor generalization to unseen degradation scenarios.

Due to their exceptional ability to model image data    distributions, diffusion models (DMs) \cite{ddpm,ncsn,score} have been successfully introduced into the field of image restoration. Their core strength lies in transforming the image generation process into a progressive denoising Markov chain, thereby implicitly encoding natural image structural priors. This characteristic enables DMs to serve as implicit regularizers, offering effective prior constraints for ill-posed inverse problems and addressing Eq.(\ref{eq2}). Currently, these methods are primarily divided into two categories: 

\begin{itemize}
\item Supervised methods require retraining models for specific degradation, which can achieve high-quality restoration but are hindered by of poor generalization and high training costs;
\item Zero-shot methods directly utilize pre-trained diffusion models through carefully designed inference mechanisms to achieve training-free image restoration, applicable to a wide range of tasks.
\end{itemize} 
Owing to their flexibility and training-free nature, zero-shot methods have gained significant attention. The existing zero-shot methods are generally classified as follows\cite{diffusionreview,diffusionreview2}:

\begin{itemize}
    \item \textbf{Projection-based Methods:}  These methods extract information from low-quality images to guide each step of the image generation process, maintaining consistency in the data. For instance, Repaint \cite{lugmayr2022repaint} employs a simple projection to add noise to the reference image and replace the known regions during each sampling step, maintaining content consistency. ILVR \cite{ilvr} constrains the generated results to align with the target image through low-frequency projection, achieving image super-resolution.
    \item \textbf{Decomposition-based method:}  These methods restore images using various decomposition strategies. DDRM \cite{ddrm} applies Singular Value Decomposition (SVD) to process the degradation matrix and solve the IR problem in the spectral domain. DDNM \cite{ddnm} ensures consistency by employing range-null space decomposition, placing observational information in range space while iteratively optimizing the null space.
    \item \textbf{Optimization-based method:}  These methods enforce consistency between each generated image and the observed image through a single-step gradient descent on the generated samples. DiffPIR \cite{diffpir} leverages the Half-Quadratic Splitting (HQS) method to alternately perform denoising and optimization, implicitly utilizing learned priors for superior image restoration. DDPG \cite{ddpg} designs a preprocessing-guided scheme that optimizes and updates the generated results at each step, resulting in more stable and refined image recovery.
    \item \textbf{Posterior estimation method:}  These methods use a pre-trained unconditional diffusion model to estimate the posterior distribution. In MCG \cite{mcg} and DPS \cite{dps}, Tweedie’s formula is used to compute the posterior mean, thereby approximating the posterior distribution. RED-Diff \cite{red-diff} and Score Prior \cite{score-prior} approximate the posterior distribution from the perspective of variational inference.
\end{itemize}

Although these methods employ different techniques, they can ultimately be reduced to the estimation of the likelihood score $\nabla_{\bm{x}_{t}} \log p(\bm{y} | \bm{x}_{t})$, as shown in Table \ref{table1}. However, because of the temporal dependency in the distribution $p(\bm{y}|\bm{x}_{t})$, obtaining an exact analytical solution for the likelihood score is difficult.

\begin{table}
\centering
\caption{ The approximation of $\nabla_{\bm{x}_{t}} \log p(\bm{y}|\bm{x}_t)$ in existing methods, ignoring the guidance strength.}
\begin{tabular*}{300pt}{@{\extracolsep\fill}ll@{\extracolsep\fill}}
\toprule
\textbf{Method} & \textbf{The approximation of} $\nabla_{\bm{x}_{t}} \log p(\bm{y}|\bm{x}_t)$\\
\midrule
DPS\cite{dps} & 
$\displaystyle \nabla_{\bm{x}_{t}} \log p(\bm{y}|\bm{x}_t) \propto -\nabla_{\bm{x}_{t}}^{T}\mathbb{E}[\bm{x}_0|\bm{x}_t] \bm{A}^{T} \left( \bm{y} - \bm{A} \mathbb{E}[\bm{x}_0|\bm{x}_t] \right)$ \\[3ex]

DDRM\cite{ddrm} &
$\displaystyle \nabla_{\bm{x}_{t}} \log p(\bm{y}|\bm{x}_t) \propto -\bm{\Sigma}^{T}|\sigma_{\bm{y}}^2\bm{I}_{m} - \sigma_{t}^2\bm{\Sigma\Sigma}^{T}|(\bar{\bm{y}}-\bm{\Sigma}\bar{\bm{x}}_{t})$\\[3ex]

\multirow{2}{*}{DDNM\cite{ddnm}} & 
$\mathbb{E}[\bm{x}_0|\bm{x}_t,\bm{y}] = (\bm{I}-\bm{A}^{\dagger}\bm{A})\mathbb{E}[\bm{x}_0|\bm{x}_t] + \bm{A}^{\dagger}\bm{y}$\\
&
$\displaystyle \nabla_{\bm{x}_{t}} \log p(\bm{y}|\bm{x}_t) \propto - \bm{A}^{\dagger}\left( \bm{y} - \bm{A} \mathbb{E}[\bm{x}_0|\bm{x}_t, \bm{y}] \right)$ \\[3ex]

\multirow{2}{*}{DiffPIR\cite{diffpir}} & 
$\displaystyle \mathbb{E}[\bm{x}_0|\bm{x}_t, \bm{y}] = \arg\min_{\bm{x}} \left\{ \frac{1}{2} \| \bm{y} - \bm{A}\bm{x} \|^2 + \frac{\lambda}{2} \| \bm{x} - \mathbb{E}[\bm{x}_0|\bm{x}_t,\bm{y}] \|^2 \right\}$ \\
& 
$\displaystyle \nabla_{\bm{x}_{t}} \log p(\bm{y}|\bm{x}_t) \propto - \bm{A}^{\dagger} \left( \bm{y} - \bm{A} \mathbb{E}[\bm{x}_0|\bm{x}_t, \bm{y}] \right)$ \\[3ex]

\multirow{3}{*}{DDPG\cite{ddpg}} & 
$\displaystyle \textbf{g}_{BP} = -\bm{A}^{T}(\bm{A}\bm{A}^{T} + \eta\bm{I}_{m})^{-1}(\bm{A}\mathbb{E}[\bm{x}_0|\bm{x}_t] - \bm{y})$ \\
&
$\displaystyle \textbf{g}_{LS} = -c\bm{A}^{T}(\bm{A}\mathbb{E}[\bm{x}_0|\bm{x}_t] - \bm{y})$ \\
&
$\displaystyle \nabla_{\bm{x}_{t}} \log p(\bm{y}|\bm{x}_t) \propto (1-\delta_{t})\textbf{g}_{BP} + \delta_{t}\textbf{g}_{LS}$ \\
\bottomrule
\end{tabular*}
\label{table1}
\end{table}

\textbf{Contribution:} To address the theoretical limitations and computational bottlenecks of existing posterior estimation methods, we propose a new guidance mechanism to efficiently tackle imaging inverse problems. The main contributions include:
\begin{itemize}
    \item First, we derive a closed-form approximation for the likelihood distribution $p(\bm{y}|\bm{x_{t}})$, along with an analytical expression for the likelihood score $\nabla _{\bm{x_{t}}}\log p(\bm{y|x_{t}})$. This avoids complex gradient backpropagation, offering a straightforward and efficient strategy to solve the inverse problem.
    \item During inference, instead of adjusting the samples directly, we guide the refinement of the model-predicted noise, making the generation process more aligned with the framework of diffusion models. Moreover, this design is inherently compatible with DDIM sampling mechanism, reducing sampling steps and further enhancing the algorithm's performance.
    \item We apply the proposed guidance mechanism to both diffusion-based sampling and iterative optimization schemes, providing flexible options to balance the different manners fo perceptual quality or accuracy.
\end{itemize}

We evaluate our method on common inverse problems using the CelebA-HQ and ImageNet datasets, demonstrating competitive performance ( Representative results are shown in Fig.\ref{fig1} ).

The organization of this paper is as follows: In Section \ref{sec2}, the fundamentals of diffusion models are briefly introduced, and the challenges associated with their application in image restoration are analyzed. In Section \ref{sec3}, an approximate analytical expression for the likelihood score is presented and we design a noise refining mechanism based on this expression. This mechanism is then applied to two generative algorithms. In Section \ref{sec4}, the effectiveness of the proposed approach is evaluated through simulation and comparative experiments on two datasets, involving various degradation case. Finally, we end the paper with a conclusion.

\section{Background and Related Work}\label{sec2}

In this section, we provide a brief overview of key concepts in diffusion models and explore how zero-shot methods leverage diffusion models to address key challenges in inverse problems of image restoration,  which motivates our research.

\subsection{Score-based Diffusion Models}\label{sec2.1}

Diffusion models, as a class of generative models, generate data via two continuous processes: forward diffusion and reverse denoising. Let $\bm{x}_{t} \in \mathbb{R}^{N}$ denote a random variable at time $t \in [0,T]$, and these two processes are mathematically described by stochastic differential equations (SDEs):

\subsubsection{Forward Diffusion Process}

The forward diffusion process gradually transforms the data distribution $p_{data}(\bm{x}_{0})$ to a Gaussian distribution by introducing noise in a controlled manner over time. This process can be modeled as a stochastic differential equation (SDE), which governs the evolution of the data as follows:
\begin{align}
    d\bm{x}_{t}=\mathit{f}(\bm{x}_{t},t)dt+\mathit{g}(t)d\bm{w}_{t},
    \label{eq3}
\end{align}
where $\mathit{f}(\bm{x}_{t},t)$ denotes the drift coefficient, representing the deterministic component of the evolution and controlling how the data point $\bm{x}_{t}$ drifts over time. $\mathit{g}(t)$ is the diffusion coefficient, which scales the noise term and controls the amount of noise added at each time step. $\bm{w}_{t}$ is the standard Wiener process (or Brownian motion), representing the stochasticity introduced during the diffusion process. When $t\xrightarrow{}T$, the system reaches the stationary distribution $\bm{x}_{T}\sim\mathcal{N}(\mathbf{0},\bm{I})$. Different choices of $\mathit{f}\,(\bm{x}_{t},t)$ and $\mathit{g}(t)$ give rise to different formulations, such as VP-SDE and VE-SDE \cite{score}.

\subsubsection{Reverse Denoising Process}

The reverse denoising process starts from a sample of standard Gaussian noise and gradually transforms it into a structured image through a sequence of denoising steps. This process is described by the time-reversed stochastic differential equation (SDE), given by
\begin{align}
    d\bm{x}_{t}=[\mathit{f}(\bm{x}_{t},t)-\mathit{g}(t)^{2}\nabla_{\bm{x}_{t}}\log p(\bm{x}_{t})]dt+\mathit{g}(t)\mathit{d}\bm{\bar{{w}}}_{t},
    \label{eq4}
\end{align}
where $\bm{\bar{w}}_{t}$ represents the time-reversed standard Wiener process, which captures the stochasticity of the reverse process. $\nabla_{\bm{x}_{t}}\log p(\bm{x}_{t})$ denotes the \textbf{score function}, representing the gradient of the log density of the data distribution. $dt$ is the negative time step, which reflects the reverse nature of the process.

The reverse denoising process essentially attempts to invert the forward diffusion process. In the forward process, noise is progressively added to the original data until until the data distribution converges to a Gaussian distribution. The reverse process seeks to undo this noise injection by progressively reducing the noise and guiding the sample back toward the true data distribution.

In the context of image restoration, the reverse denoising process plays a central role in recovering corrupted images. By carefully modeling the score function and optimizing the reverse trajectory, it becomes possible to reconstruct high-quality images from degraded measurements. 

\subsubsection{Score Matching}

A key challenge in the reverse process is estimating the score function $\nabla_{\bm{x}_{t}}\log p(\bm{x}_{t})$ which is generally intractable. The unknown score function is approximated by a neural network trained via denoising score matching \cite{dsm}:
\begin{align}
    \bm{\theta ^{*}} = \arg\min_{\bm{\theta}} \, \mathbb{E}_{t} \Bigl\{ 
    \lambda (t) \mathbb{E}_{\bm{x}_{0}} \mathbb{E}_{\bm{x}_{t}|\bm{x}_{0}} 
    \bigl[ & ||\bm{s_{\theta}}(\bm{x}_{t},t)  - \nabla _{\bm{x}_{t}} \log p_{t|0}(\bm{x}_{t}|\bm{x}_{0})||_{2}^{2} \bigr] 
    \Bigr\},
    \label{eq5}
\end{align}
where $\lambda(t)$ represents the weighting coefficient, $\bm{x}_{0}\sim p_{\mathrm{data}}(\bm{x}), \bm{x}_{t}\sim p_{t|0}(\bm{x}_{t}|\bm{x}_{0})$. 

Once $\bm{s_{\theta}}(\bm{x}_{t},t)$ is obtained from Eq.(\ref{eq5}), the score function in Eq.(\ref{eq4}) can be replaced using the approximation $\nabla_{\bm{x}_{t}}\log p(\bm{x}_{t}) \simeq \bm{s_{\theta}}(\bm{x}_{t},t)$. Subsequently, solving Eq.(\ref{eq4}) enables the generation of samples.

\subsection{Denoising Diffusion Probabilistic Model}\label{sex2.2}

The diffusion model adopted in this paper is the Denoising Diffusion Probabilistic Model (DDPM) \cite{ddpm}. DDPM employs fixed variance schedules $\left\{ \beta_{t} \right\}_{t=1}^{T}$ for forward corruption:
\begin{align}
    \bm{x}_{t}=\sqrt{1-\beta_{t}}\bm{x}_{t-1}+\sqrt{\beta_{t}}\bm{\epsilon}_{t-1},
    \label{eq6}
\end{align}
where $\bm{\epsilon}_{t-1} \sim \mathcal{N}(\bm{0},\bm{I})$ represents the noise sampled randomly at each step. As $T\to \infty$, Eq.(\ref{eq6}) converges to the following VP-SDE\cite{score}:
\begin{align}
    d\bm{x}_{t}=-\frac{1}{2}\beta(t)dt+\sqrt{\beta(t)} d\bm{w}_{t}.
    \label{eq7}
\end{align}
Therefore, DDPM can be viewed as a discrete formulation of the VP-SDE.

Based on the properties of the Gaussian distribution, the noised result at any time step $\mathit{t}$ can be derived as:
\begin{align}
\bm{x}_{t}=\sqrt{\bar{\alpha}_{t}}\bm{x}_{0}+\sqrt{1-\bar{\alpha}_{t}}\bm{\epsilon},
\label{eq8}
\end{align}
where $\bar{\alpha}_{t}=\textstyle \prod_{i=1}^{t}\alpha _{i},\alpha_{t}=1-\beta_{t}$. Then the reverse diffusion process formula for each step in DDPM is as follows:
\begin{align}
    \bm{x}_{t-1}=\frac{1}{\sqrt{\alpha_{t}}}\left(\bm{x}_{t}-\frac{\beta_{t}}{\sqrt{1-\bar{\alpha}_{t}}}\bm{\epsilon}_{\theta}(\bm{x}_{t},t)\right)+\sqrt{\beta_{t}}\bm{\epsilon}_{t},
    \label{eq9}
\end{align}
$\bm{\epsilon}_{\theta}(\bm{x}_{t},t)$ is the noise predictor, which is used to predict the noise component at each time step $t$. The relationship between the score function $\nabla_{\bm{x}_{t}}\log p(\bm{x}_{t})$ and noise prediction $\bm{\epsilon}_{\theta}(\bm{x}_{t},t)$ can be approximated as:
\begin{align}
    \nabla_{\bm{x}_{t}}\log p(\bm{x}_{t})\approx
    -\frac{1}{\sqrt{1-\bar{\alpha}_{t}}}\bm{\epsilon}_{\theta}(\bm{x}_{t},t).
    \label{eq10}
\end{align}

Although DDPM demonstrates strong generative capabilities, its inference efficiency is significantly limited by multi-step sampling. To address this issue, an efficient sampling method known as Denoising Diffusion Implicit Models (DDIM) \cite{ddim} has been proposed. DDIM extends the diffusion process from a Markovian to a non-Markovian process, enabling a significant reduction in sampling steps while maintaining generation quality. The sampling formula is rewritten as:
\begin{align}
    \bm{x}_{t-1}=\sqrt{\bar{\alpha}_{t-1}}\left(\frac{\bm{x_{t}} - \sqrt{1-\bar{\alpha}_{t}} \bm{\epsilon}_{\bm{\theta}}(\bm{x_{t}},t)}{\sqrt{\bar{\alpha}_{t}}}\right)+\sqrt{1-\bar{\alpha}_{t-1}-\sigma^{2}}\cdot\bm{\epsilon}_{\mathbf{\theta}}(\bm{x}_{t},t)+\sigma\bm{\epsilon_{t}},
    \label{eq11}
\end{align}
where $\bm{\epsilon_{t}}$ denotes standard Gaussian noise, $\sigma$ controls the weight between the stochastic noise $\bm{\epsilon_{t}}$ and the deterministic predicted noise $\bm{\epsilon}_{\bm{\theta}}(\bm{x}_{t},t)$.

\subsection{Solving Linear Imaging Inverse Problems With Diffusion Models}\label{sec2.3}

Using diffusion models to restore degraded images is equivalent to solving a conditional generation problem, i.e., sampling from the conditional distribution $p(\bm{x}_{t}|\bm{y})$ to generate the target image. In this framework, the score function $\nabla_{\bm{x}_{t}}\log p(\bm{x}_{t})$ transitions from an unconditional score to a conditional score $\nabla_{\bm{x}_{t}}\log p(\bm{x}_{t}|\bm{y})$. The conditional score $\nabla_{\bm{x}_{t}}\log p(\bm{x}_{t}|\bm{y})$ can be decomposed using Bayes' theorem as follows:
\begin{align}
    \underbrace{\nabla_{\bm{x}_{t}}\log p(\bm{x}_{t}|\bm{y})}_{\textit{conditional score}} = \underbrace{\nabla_{\bm{x}_{t}}\log p(\bm{x}_{t})}_{\textit{unconditional score}} + \underbrace{\nabla_{\bm{x}_{t}}\log p(\bm{y}|\bm{x}_{t})}_{\textit{likelihood score}}.
    \label{eq12}
\end{align}

It can be seen that the conditional score is divided into two parts: the unconditional score and the likelihood score. Therefore, the reverse SDE Eq.(\ref{eq4}) becomes:
\begin{align}
    d\bm{x}_{t} = & [f(\bm{x}_{t},t) - g(t)^{2} (\nabla_{\bm{x}_{t}} \log p(\bm{x}_{t})  + \nabla_{\bm{x}_{t}} \log p(\bm{y} | \bm{x}_{t}))] dt + \mathit{g}(t) d \bar{\bm{w}}_{t}.
    \label{eq13}
\end{align}

The unconditional score corresponds to the pre-trained diffusion model, while the core issue lies in obtaining the likelihood score. However, due to its strong dependence on the time step $t$, the calculation of $\nabla_{\bm{x}_{t}} \log p(\bm{y} | \bm{x}_{t})$ becomes highly challenging.

\vspace{10pt}
\textit{(\textbf{Tweedie's formula}\cite{dps})  For the case of DDPM sampling, $p(\bm{x}_{0}|\bm{x}_{t})$ has the unique posterior mean at:
\begin{align}
    \bm{x}_{0|t} & = \frac{1}{\sqrt{\bar{\alpha}_{t}}}(\bm{x}_{t} + (1-\bar{\alpha}_{t})\nabla_{\bm{x}_{t}}\log p(\bm{x}_{t})).
    \label{eq14}
\end{align}}

By replacing the $\nabla_{\bm{x}_{t}}\log p(\bm{x}_{t})$ with the approximation in Eq.(\ref{eq10}), the posterior mean of $p(\bm{x}_{0}|\bm{x}_{t})$ can be approximated as:
\begin{align}
     \bm{x}_{0|t}\approx \frac{1}{\sqrt{\bar{\alpha}_{t}}}(\bm{x}_{t} - \sqrt{1-\bar{\alpha}_{t}}\bm{\epsilon}_{\theta}(\bm{x}_{t},t)).
     \label{eq15}
\end{align}

DPS\cite{dps} employs the Laplace approximation and utilizes Tweedie’s approach\cite{tweedie,kim2021noise2score} to approximate the distribution $p(\bm{y}|\bm{x}_{t})$ with $p(\bm{y}|\bm{x}_{0|t})$. The gradient of the likelihood term $\nabla_{\bm{x}_{t}}\log p(\bm{y}|\bm{x}_{t})$ is ultimately approximated as:
\begin{align}
    \nabla_{\bm{x}_{t}}\log p(\bm{y}|\bm{x}_{t}) = -\rho\,\nabla _{\bm{x}_{t} }\left \| \bm{y}-\bm{A}\bm{x}_{0|t} \right \| _{2}^{2}.
    \label{eq16}
\end{align}
Here, $\rho$ represents the guidance strength, and the gradient term can be computed via backpropagation. However, this approximation method essentially amounts to a form of point estimation, neglecting the variance properties of the distribution. Furthermore, the recovery process updates the generated result at each step using Eq.(\ref{eq16}). The lengthy denoising steps, combined with gradient backpropagation computations, significantly reduce the algorithm's efficiency and limit its practical applicability.

\begin{figure}[t]
    \centering
    \includegraphics[width=0.9\textwidth]{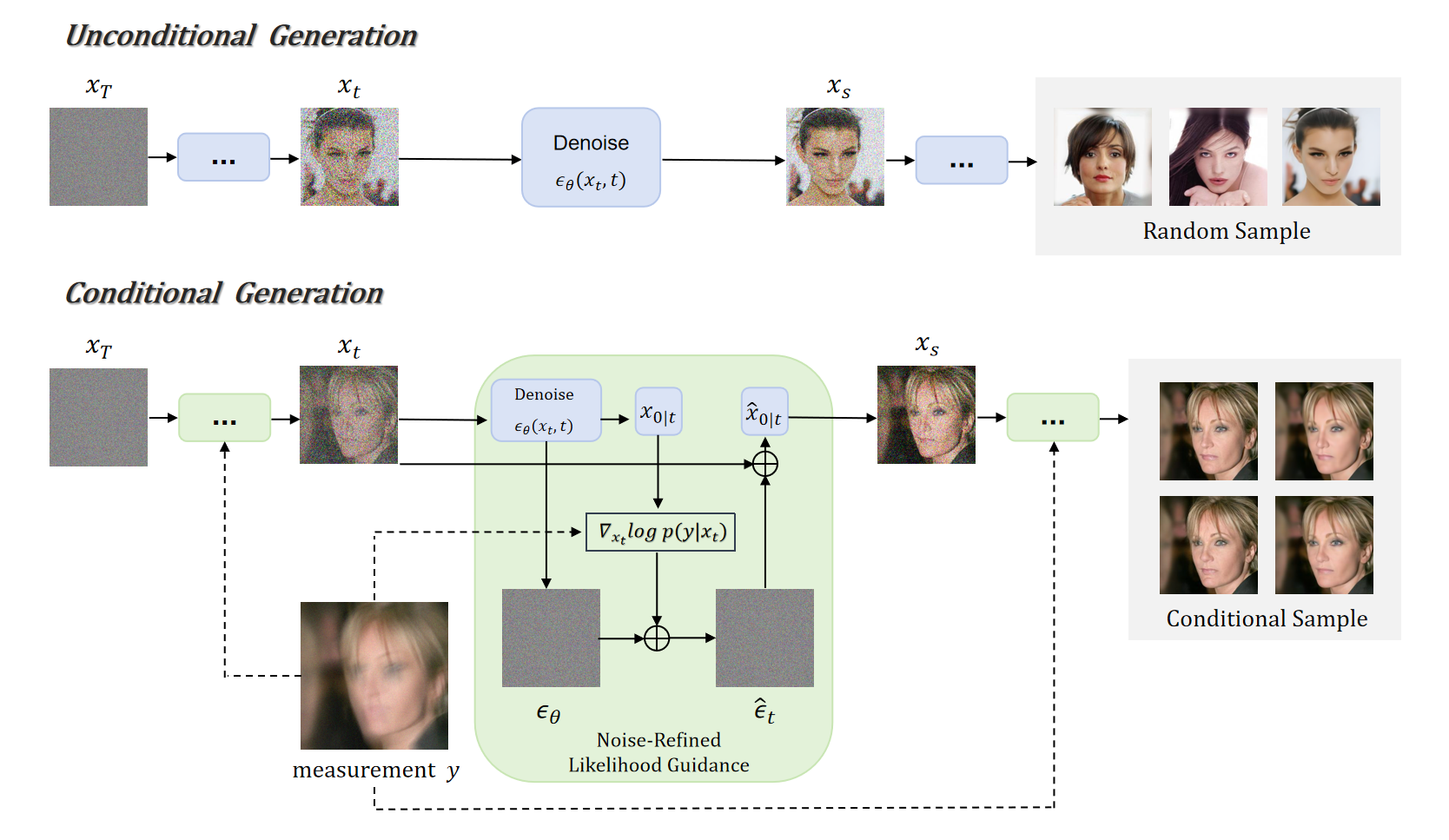} 
    \caption{\textbf{Illustration of sampling method}. (Top) Unconditional generation with pretrained diffusion models. (Bottom) Our approach leverages pretrained diffusion models for conditional generation. For every state $\bm{x}_t$, using the measurement $\bm{y}$ and the predicted $\bm{x}_{0|t}$
    from the diffusion model to construct a likelihood score $\nabla_{\bm{x}_{t}}\log p(\bm{y}|\bm{x}_{t})$, and the unconditional noise $\bm{\epsilon}_\theta$ is refined to a conditional noise $\hat{\bm{\epsilon}}_t$. Subsequently, the next state $\bm{x}_{t-1}$ is derived by adding noise back to new $\hat{\bm{x}}_{0|t}$. }
  \label{fig2} 
\end{figure}

\section{Proposed Method}\label{sec3}

In this section, we propose an effective approximation of the likelihood score $\nabla_{\bm{x}_{t}}\log p(\bm{y}|\bm{x}_{t})$. Unlike previous approaches that directly adjust generated samples, our method implicitly enforces prior constraints by refining directions in the noise prediction space. This not only aligns with the generative framework of diffusion models, but is also naturally compatible with DDIM-accelerated sampling. Furthermore, we design two schemes—one sampling-based and the other on iterative optimization—offering flexible and efficient solutions for solving inverse problems.

\subsection{Likelihood Score Approximation}\label{sec3.1}

As mentioned in Section \ref{sec2.3}, the core issue in achieving image restoration using diffusion models is to estimate the likelihood score $\nabla _{\bm{x}_{t}}\log p(\bm{y}|\bm{x}_{t})$ in Eq.(\ref{eq13}).To utilize the degradation model in Eq.(\ref{eq1}), the likelihood distribution can be decomposed as follows:
\begin{align}
    p(\bm{y}|\bm{x}_{t})=\int p(\bm{y}|\bm{x}_{0})p(\bm{x}_{0}|\bm{x}_{t})d\bm{x}_{0},
    \label{eq17}
\end{align}
where $p(\bm{y}|\bm{x}_{0})=\mathcal{N}(\bm{Ax}_{0},\sigma_{\bm{y}}^{2}\bm{I})$ is determined by the degradation model. However, the reverse transition probability $p(\bm{x}_{0}|\bm{x}_{t})$ is difficult to obtain directly, leading to an inexact representation of the distribution $p(\bm{y}|\bm{x}_{t})$.

Next, we derive the expression for $p(\bm{x}_{0}|\bm{x}_{t})$. Recalling the forward process in Eq.(\ref{eq8}), we can easily obtain 
\begin{align}
    \bm{x}_0 = \frac{1}{\sqrt{\bar{\alpha}_{t}}}\bm{x}_{t}-\sqrt{\frac{1-\bar{\alpha}_{t}}{\bar{\alpha}_{t}}}\bm{\epsilon}.
    \label{eq18}
\end{align}
Therefore, an approximation for $p(\bm{x}_{0}|\bm{x}_{t})$ can be derived as:
\begin{align}
    p(\bm{x}_{0}|\bm{x}_{t})\simeq
    \mathcal{N}\left(\frac{1}{\sqrt{\bar{\alpha}_{t}}}\bm{x}_{t},\frac{1-\bar{\alpha}_{t}}{\bar{\alpha}_{t}}\bm{I}\right).
    \label{eq19}
\end{align}

However, the above derivation of $p(\bm{x}_{0}|\bm{x}_{t})$ relies solely on the forward diffusion process in Eq.(\ref{eq8}). By applying Bayes' theorem to $p(\bm{x}_{0}|\bm{x}_{t})$, we can obtain:
\begin{align}
    p(\bm{x}_{0}|\bm{x}_{t})=
    \frac{p(\bm{x}_{t}|\bm{x}_{0})p(\bm{x}_{0})}{p(\bm{x}_{t})}.
    \label{eq20}
\end{align}
We observe that $p(\bm{x}_{0}|\bm{x}_{t})\propto
p(\bm{x}_{t}|\bm{x}_{0})p(\bm{x}_{0})$, indicating that $p(\bm{x}_{0}|\bm{x}_{t})$ is not only determined by the forward diffusion process but also incorporates the prior information $p(\bm{x}_{0})$. However, in Eq.(\ref{eq19}), the mean of the distribution $p(\bm{x}_{0}|\bm{x}_{t})$ depends solely on $\bm{x}_{t}$ which is unreasonable. Therefore, it is necessary to revise this distribution to ensure that it incorporates the prior information associated with $\bm{x}_{0}$.

Naturally, we turn to Tweedie's formula in Eq.(\ref{eq15}). The posterior mean computed by Tweedie's formula corresponds to the optimal posterior mean (The proof of optimality is provided in Appendix \ref{App.A.1}) under the Minimum Mean Square Error (MMSE) criterion. Specifically:
\begin{align}
    \bm{x}_{0|t}=\mathbb{E}[\bm{x}_0|\bm{x}_t]\approx \frac{1}{\sqrt{\bar{\alpha}_{t}}}(\bm{x}_{t}-\sqrt{1-\bar{\alpha}_{t}}\bm{\epsilon}_{\theta}(\bm{x}_{t},t))
    \label{eq21}
\end{align}
This formula not only provides the optimal posterior mean but also computes it using the learned noise predictor $\bm{\epsilon}_{\theta}(\bm{x}_{t},t)$ which implicitly captures the prior distribution of the data $\bm{x}_{0}$.  This further validating the rationale behind the mean correction.

Therefore, we perform a mean correction in Eq.(\ref{eq19}) by replacing the original posterior mean with the one obtained from Tweedie’s formula. Consequently, the distribution $p(\bm{x}_{0}|\bm{x}_{t})$ is corrected as:
\begin{align}
    p(\bm{x}_{0}|\bm{x}_{t}) \simeq
    \mathcal{N}\left(\frac{1}{\sqrt{\bar{\alpha}_{t}}}(\bm{x}_{t}-\sqrt{1-\bar{\alpha}_{t}}\bm{\epsilon}_{\theta}(\bm{x}_{t},t)),\frac{1-\bar{\alpha}_{t}}{\bar{\alpha}_{t}}\bm{I}\right).
    \label{eq22}
\end{align}
 Then, to derive the $p(\bm{y}|\bm{x}_{t})$, we give the following property.

\vspace{5pt}
\textit{\textbf{Property:}} \textit{If the random variables $\bm{z}_{0},\bm{z}_{1},\bm{z}_{2}$ satisfy the following conditional probability distribution $ p(\bm{z}_{1} | \bm{z}_{0})=\mathcal{N}(\bm{z}_{0}, V_1) $, $ p(\bm{z}_{2} | \bm{z}_{1}) = \mathcal{N}(\alpha \bm{z}_{1}, V_2) $, then $ p(\bm{z}_{2} | \bm{z}_{0})$ satisfy the following distribution (proof in Appendix \ref{APP.A.2}):}
\[
p(\bm{z}_{2} | \bm{z}_{0})= \mathcal{N}(\alpha \bm{z}_{0}, \alpha^2 V_1 + V_2).
\]

By combining the Eq.(\ref{eq1}), Eq.(\ref{eq22}) and Property, we can obtain the likelihood distribution as follows:
\begin{align}
    p(\bm{y}|\bm{x}_{t})\simeq\mathcal{N}\left(\bm{A}\cdot\frac{1}{\sqrt{\bar{\alpha}_{t}}}(\bm{x}_{t}-\sqrt{1-\bar{\alpha}_{t}}\bm{\epsilon}_{\theta}(\bm{x}_{t},t)),\frac{1-\bar{\alpha}_{t}}{\bar{\alpha}_{t}}\bm{AA}^{T}+\sigma_{\bm{y}}^{2}\bm{I}\right).
    \label{eq23}
\end{align}
Thus, through a straightforward derivation of the log-gradient, the analytical expression for the likelihood score can be expressed as:
\begin{align}
    \nabla_{\bm{x}_{t}} \log p(\bm{y}|\bm{x}_{t}) \simeq
    -\frac{1}{\sqrt{\bar{\alpha}_{t}}} \bm{A}^{T} \left( \frac{1-\bar{\alpha}_{t}}{\bar{\alpha}_{t}} \bm{A} \bm{A}^{T} + \sigma_{\bm{y}}^{2} \bm{I} \right)^{-1} 
    \cdot \left( \bm{y} - \bm{A}\bm{x}_{0|t} \right)\cdot(\bm{I}-\sqrt{1-\bar{\alpha}_{t}}\nabla_{\bm{x}_{t}}\bm{\epsilon}_{\theta}(\bm{x}_{t},t)),
    \label{eq24}
\end{align}
where $\bm{x}_{0|t}=\frac{1}{\sqrt{\bar{\alpha}_{t}}}(\bm{x}_{t}-\sqrt{1-\bar{\alpha}_{t}}\bm{\epsilon}_{\theta}(\bm{x}_{t},t))$. 

In Eq. (\ref{eq24}), the last multiplier term is the gradient of the pretrained network's noise with respect to $\bm{x}_{t}$, which involves considerable computation Jacobian matrix. However, in the forward diffusion process defined by Eq. (\ref{eq8}), the injected noise at each diffusion step is independently sampled from a Gaussian distribution that remains uncorrelated with the current state. Motivated by this property, we introduce the following assumption:

\vspace{5pt}
\textit{\textbf{Assumption:}}\textit{The noise prediction network output $\bm{\epsilon}_{\theta}(\bm{x}_{t},t)$ is assumed to be independent of the input variable $\bm{x}_t$. Formally, this implies that the Jacobian matrix of $\bm{\epsilon}_{\theta}(\bm{x}_{t},t)$ with respect to $\bm{x}_t$ vanishes $\nabla_{\bm{x}_{t}}\bm{\epsilon}_{\theta}(\bm{x}_{t},t)=0$.}

\vspace{5pt}
Based on this assumption, we can omit the last term in Eq. (\ref{eq24}), leading to the following approximation of the likelihood score:

\begin{align}
    \nabla_{\bm{x}_{t}} \log p(\bm{y}|\bm{x}_{t}) \simeq
    -\frac{1}{\sqrt{\bar{\alpha}_{t}}} \bm{A}^{T} \left( \frac{1-\bar{\alpha}_{t}}{\bar{\alpha}_{t}} \bm{A} \bm{A}^{T} + \sigma_{\bm{y}}^{2} \bm{I} \right)^{-1} 
    \cdot \left( \bm{y} - \bm{A}\bm{x}_{0|t} \right).
    \label{eq25}
\end{align}

After obtaining the analytical expression of the likelihood score Eq.(\ref{eq25}), image restoration can be achieved using Eq.(\ref{eq13}). Specifically, let $\bm{A} = \bm{U}\bm{\Sigma}\bm{V}^{T}$. Then, the likelihood score can be efficiently computed using the following Singular Value Decomposition (SVD) method.
\begin{align}
    \nabla_{\bm{x}_{t}} \log p(\bm{y}|\bm{x}_{t}) \simeq - \frac{1}{\sqrt{\bar{\alpha}_{t}}}\bm{V}\bm{\Sigma}\left(\frac{1-\bar{\alpha}_{t}}{\bar{\alpha}_{t}}\bm{\Sigma}^{2} + \sigma_{\bm{y}}^{2} \bm{I} \right)^{-1}\bm{U}^{T}( \bm{y} - \bm{A}\bm{x}_{0|t}),
    \label{eq26}
\end{align}
where $\bm{\Sigma}^2$ denotes elements-wise square of $\bm{\Sigma}$.

Correspondingly, $\Pi$GDM \cite{PGDM} also proposes a similar approximation form of the likelihood score. However, in modeling the distribution $p(\bm{x}_0|\bm{x}_t)$, it directly simplifies the posterior variance as a tunable hyperparameter, whereas in our work, the variance information is theoretically derived based on the formulation of the forward diffusion process. In addition, the method in $\Pi$GDM still relies on the explicit computation of the Jacobian matrix, which incurs a substantial computational overhead in practical applications. Furthermore, regarding the application of the likelihood score, $\Pi$GDM applies it directly to the update of sample states. In contrast, we argue that noise plays a critical role in determining the generative quality of diffusion models. Therefore, we introduce a likelihood-score-based noise refinement strategy, in which the likelihood score is incorporated into the update of the predicted noise, as detailed in Section \ref{sec3.2}.

\subsection{Noise Refinement Strategy}\label{sec3.2}

In diffusion models, images are generated by progressively denoising at each step, and the accuracy of the noise prediction directly determines the quality of the generation. Therefore, we aim to achieve the restoration by refining the noise, rather than directly adjusting the samples as in previous practices. However, the noise predicted by the pre-trained diffusion model is unconditional, which cannot effectively guide the model to the true distribution $p(\bm{x}_{0}|\bm{y})$ of the data to be restored. Thus, we introduce the likelihood score to guide the refinement of the pre-predicted noise at each step, adjusting the denoising direction to make the generated result close to the true data distribution.

In DDPM, the relationship between the score function and noise prediction can be approximated by Eq.(\ref{eq10}). For the condition score function $\nabla_{\bm{x}_{t}}\log p(\bm{x}_{t}|\bm{y}) $ in Eq.(\ref{eq12}), by substituting the score function with the predicted noise as that in Eq.(\ref{eq10}), it can be rewritten as follows: 
\begin{align}
    \nabla_{\bm{x_{t}}}\log p(\bm{x}_{t}|\bm{y}) &= -\frac{1}{\sqrt{1-\bar{\alpha}_{t}}}\bm{\epsilon}_{\theta}(\bm{x}_{t},t) + \nabla_{\bm{x}_{t}}\log p(\bm{y}|\bm{x}_{t})\nonumber\\
    &=-\frac{1}{\sqrt{1-\bar{\alpha}_{t}}}(\bm{\epsilon}_{\theta}(\bm{x}_{t},t)-\sqrt{1-\bar{\alpha}_{t}}\nabla_{\bm{x}_{t}}\log p(\bm{y}|\bm{x}_{t}))\nonumber\\
    &=-\frac{1}{\sqrt{1-\bar{\alpha}_{t}}} \hat{\bm{\epsilon}}_{t}(\bm{x}_{t},t),
    \label{eq27}
\end{align}
where  $\hat{\bm{\epsilon}}_{t}(\bm{x}_{t},t)= \bm{\epsilon}_{\theta}(\bm{x}_{t},t)-\sqrt{1-\bar{\alpha}_{t}}\nabla_{\bm{x}_{t}}\log p(\bm{y}|\bm{x}_{t})$.

By comparing Eq.(\ref{eq10}) and Eq.(\ref{eq27}), we observe that Eq.(\ref{eq27}) provides the relationship between the conditional score function and noise $\hat{\bm{\epsilon}}_{t}(\bm{x}_{t},t)$, in which the noise is based on a single gradient descent to the pre-trained noise $\bm{\epsilon}_{\theta}(\bm{x}_{t},t)$. Generally, by introducing a step $\mu$ in gradient descent, we can refine the noise as follows:
\begin{align}
    \hat{\bm{\epsilon}}_{t}(\bm{x}_{t},t)=\bm{\epsilon}_{\theta}(\bm{x}_{t},t)-\mu\sqrt{1-\bar{\alpha}_{t}}\nabla_{\bm{x}_{t}}\log p(\bm{y}|\bm{x}_{t}).
    \label{eq28}
\end{align}

The proposed noise refinement strategy plays a pivotal role in integrating condition-guided information into the noise prediction process, thereby transforming unconditional noise $\bm{\epsilon}_{\theta}(\bm{x}_{t},t)$ into conditional noise $\hat{\bm{\epsilon}}_{t}(\bm{x}_{t},t)$ and guiding the generated data toward the true data distribution. This aligns perfectly with the fundamental generative principle of the diffusion models.

\subsection{Sampling Scheme}\label{sec3.3}

To improve sampling efficiency, we employ the DDIM sampling method instead of the original DDPM sampling. Within the DDIM framework, as discussed in Section \ref{sec3.2}, after refining the model-predicted noise to obtain $\hat{\bm{\epsilon}}_{t}(\bm{x}_{t},t)$, we get a one-step sampling formula similar to Eq.(\ref{eq11}):
\begin{align}
    \bm{x}_{t-1}=\sqrt{\bar{\alpha}_{t-1}}\left(\frac{\bm{x_{t}} - \sqrt{1-\bar{\alpha}_{t}} \hat{\bm{\epsilon}}_{t}(\bm{x}_{t},t)}{\sqrt{\bar{\alpha}_{t}}}\right)+\sqrt{1-\bar{\alpha}_{t-1}-\sigma^{2}}\cdot\hat{\bm{\epsilon}}_{t}(\bm{x}_{t},t)+\sigma\bm{\epsilon_{t}},
    \label{eq29}
\end{align}
where $\bm{\epsilon}_{t} \sim \mathcal{N}(\bm{0},\bm{I})$ denotes the stochastic noise.

Furthermore, to simplify the form of noise injection, we introduce a parameter  $\mathbf{\zeta}\in[0,1]$ as weight to adjust the deterministic noise $\hat{\bm{\epsilon}}_{t}(\bm{x}_{t},t)$ and the stochastic noise $\bm{\epsilon}_{t}$. Specifically, the noise injection in Eq.(\ref{eq29}) is replaced by:
\begin{align}
    \bm{x}_{t-1}=\sqrt{\bar{\alpha}_{t-1}}\hat{\bm{x}}_{0|t}+\sqrt{1-\bar{\alpha}_{t-1}}\left(\sqrt{1-\zeta}\hat{\bm{\epsilon}_{t}}(\bm{x}_{t},t)+\sqrt{\zeta}\bm{\epsilon}_{t}\right),
    \label{eq30}
\end{align}
where $\hat{\bm{x}}_{0|t}=\frac{\bm{x}_{t}-\sqrt{1-\bar{\alpha}_{t}}\hat{\bm{\epsilon}}_{t}(\bm{x}_{t},t)}{\sqrt{\bar{\alpha}_{t}}}$  represents the refined denoised estimate. When $\mathbf{\zeta}=0$, the sampling process is fully deterministic, corresponding to standard DDIM. Conversely, when $\mathbf{\zeta}=1$, the process incorporates full stochasticity, allowing for diverse sample generation.

We term this method Denoising Diffusion with Noise-Refined Likelihood Guidance (DD-NRLG), as detailed in Algorithm \ref{alg1}, and an illustration of the overall workflow is shown in Figure \ref{fig2}. Notably, our scheme only requires the adjustment of two key parameters: the noise refinement strength $\mu$ and the noise injection parameter $\zeta$, which ensures both flexibility and ease of implementation.

\begin{figure}[t]
    \centering
    \begin{minipage}[t]{0.48\linewidth}
        \renewcommand{\thealgorithm}{1}
        \begin{algorithm}[H]
            \footnotesize  
            \caption{Denoising Diffusion with Noise-Refined Likelihood Guidance}
            \label{alg1}
            \begin{algorithmic}[1]
                \setlength{\itemsep}{0.5em}
                \Require $\bm{\epsilon}_{\theta}(\cdot,t)$, T, $\bm{y}$, $\sigma_{\bm{y}}$, $\bm{A}$, $\left\{\bar{\alpha}_{t}\right\}$, $\mu$, $\zeta$
                \State $\mathrm{Initialize}\;\bm{x}_{T}\sim\mathcal{N}(\bm{0},\bm{I}_{n})$
                \For{$t$ from $T$ to $1$}
                    \State $\bm{x}_{0|t}= \frac{1}{\sqrt{\bar{\alpha}_{t}}}\left(\bm{x}_{t} - \sqrt{1-\bar{\alpha}_{t}} \bm{\epsilon}_{\bm{\theta}}(\bm{x}_{t},t)\right)$
                    \State $\nabla_{\bm{x}_{t}} \log p(\bm{y}|\bm{x}_{t}) \simeq -\frac{1}{\sqrt{\bar{\alpha}_{t}}} \bm{A}^{T} \left( \frac{1-\bar{\alpha}_{t}}{\bar{\alpha}_{t}} \bm{A} \bm{A}^{T} + \sigma_{\bm{y}}^{2} \bm{I} \right)^{-1} \left( \bm{y} - \bm{A}\bm{x}_{0|t} \right)$
                    \State $\hat{\bm{\epsilon}}_{t}(\bm{x}_{t},t)=\bm{\epsilon}_{\theta}(\bm{x}_{t},t)-\mu\sqrt{1-\bar{\alpha}_{t}}\nabla_{\bm{x}_{t}}\log p(\bm{y}|\bm{x}_{t})$
                    \State $\hat{\bm{x}}_{0|t}=\frac{1}{\sqrt{\bar{\alpha}_{t}}}\left(\bm{x}_{t}-\sqrt{1-\bar{\alpha}_{t}}\hat{\bm{\epsilon}}_{t}(\bm{x}_{t},t)\right)$
                    \State $\bm{\epsilon}_{t} \sim \mathcal{N}(\bm{0}, \bm{I}_n)$
                    \State $\bm{x}_{t-1}=\sqrt{\bar{\alpha}_{t-1}}\hat{\bm{x}}_{0|t}+\sqrt{1-\bar{\alpha}_{t-1}}\left(\sqrt{1-\zeta}\,\hat{\bm{\epsilon}}_{t}(\bm{x}_{t},t)+\sqrt{\zeta}\,\bm{\epsilon}_{t}\right)$
                \EndFor
                \State \Return $\bm{x}_0$
            \end{algorithmic}
        \end{algorithm}
    \end{minipage}
    \hspace{0.02\linewidth}
    \begin{minipage}[t]{0.48\linewidth}
        \renewcommand{\thealgorithm}{2}
        \begin{algorithm}[H]
            \footnotesize  
            \caption{Iterative Denoising with Noise-Refined Likelihood Guidance}
            \label{alg2}
            \begin{algorithmic}[1]
                \setlength{\itemsep}{0.5em}
                \Require $\bm{\epsilon}_{\theta}(\cdot,t)$, T, $\bm{y}$, $\sigma_{\bm{y}}$, $\bm{A}$, $\left\{\bar{\alpha}_{t}\right\}$, $\mu$
                \State $\mathrm{Initialize}\;\bm{x}_{T}\sim\mathcal{N}(\bm{0},\bm{I}_{n})$
                \For{$t$ from $T$ to $1$}
                    \State $\bm{x}_{0|t}= \frac{1}{\sqrt{\bar{\alpha}_{t}}}\left(\bm{x}_{t} - \sqrt{1-\bar{\alpha}_{t}} \bm{\epsilon}_{\bm{\theta}}(\bm{x}_{t},t)\right)$
                    \State $\nabla_{\bm{x}_{t}} \log p(\bm{y}|\bm{x}_{t}) \simeq -\frac{1}{\sqrt{\bar{\alpha}_{t}}} \bm{A}^{T} \left( \frac{1-\bar{\alpha}_{t}}{\bar{\alpha}_{t}} \bm{A} \bm{A}^{T} + \sigma_{\bm{y}}^{2} \bm{I} \right)^{-1} \left( \bm{y} - \bm{A}\bm{x}_{0|t} \right)$
                    \State $\hat{\bm{\epsilon}}_{t}(\bm{x}_{t},t)=\bm{\epsilon}_{\theta}(\bm{x}_{t},t)-\mu\sqrt{1-\bar{\alpha}_{t}}\nabla_{\bm{x}_{t}}\log p(\bm{y}|\bm{x}_{t})$
                    \State $\hat{\bm{x}}_{0|t}=\frac{1}{\sqrt{\bar{\alpha}_{t}}}\left(\bm{x}_{t}-\sqrt{1-\bar{\alpha}_{t}}\hat{\bm{\epsilon}}_{t}(\bm{x}_{t},t)\right)$
                    \State $\bm{x}_{t-1}=\sqrt{\bar{\alpha}_{t-1}}\hat{\bm{x}}_{0|t}$
                \EndFor
                \State \Return $\bm{x}_0$
            \end{algorithmic}
        \end{algorithm}
    \end{minipage}
\end{figure}

\subsection{Iterative Optimization Scheme}\label{3.4}

In this section, we further extend our guiding strategy to deterministic optimization via the Plug-and-Play (PnP) framework. 

The traditional model-based image restoration approach is to solve the optimization problem in Eq.(\ref{eq2}). However, handcrafted priors exhibit limited expressiveness for complex data structures. Instead of explicitly defining the regularization term, the Plug-and-Play (PnP) framework \cite{plug1,plug2} leverages implicit priors learned from pre-trained denoising models, allowing flexible integration of advanced denoising algorithms into the iterative optimization process. Based on the PnP framework, image restoration can be achieved through iterative optimization, in which the data consistency is dictated by the physical degradation model, and the denoiser serves as a projection operation for the implicit prior. 

Since PnP provides a flexible framework for combining state-of-the-art denoiser as a prior, diffusion models can be integrated into the PnP framework\cite{diffpir,ddpg,plugdiffusion}. Specifically, the prior is replaced by a pre-trained  denoiser $\mathcal{D}(\cdot,\sigma_{t}), $(where $\sigma_{t}$ represents the noise level), and the iterative image restoration can be formulated as:
\begin{align}
    &\bm{x}_{0|t}=\mathcal{D}(\bm{x}_{t},\sigma_{t}), \label{eq31}\\
    &\bm{x}_{t-1}=\bm{x}_{0|t}-\mu_{t}\nabla_{\bm{x}}\ell(\bm{x}_{0|t},\bm{y}).
    \label{eq32}
\end{align}

Under this iterative optimization scheme, the generated samples are first denoised through the diffusion process, followed by an update step via gradient descent. It can be observed that there is no noise re-injection step (step 8 in Algorithm \ref{alg1}) throughout the entire iterative framework. A reasonable explanation is that the optimization-based method aims to find a unique solution that not only aligns with the observed data but also satisfies the inherent prior information. In contrast, reintroducing random noise introduces stochasticity into the generation process, leading to a set of possible solutions that conform to a certain data distribution. 

Thus, we deliberately remove the noise injection step to enforce determinism in the optimization process, ensuring the recovery of a unique optimal solution. The proposed optimization scheme is summarized in Algorithm \ref{alg2}. In particular, the noise refinement strategy proposed in Eq.(\ref{eq28}) implicitly facilitates the overall process in performing gradient descent in Eq.(\ref{eq32}).

 Comparing the two schemes, Algorithm \ref{alg2} eliminates noise re-injection, thereby achieving a unique optimal solution and ensuring high data fidelity. In contrast, Algorithm \ref{alg1} leverages stochastic noise injection, aligning more closely with the generative paradigm of generative models, and produces more details. While Algorithm \ref{alg1} may compromise data fidelity, it could outperform Algorithm \ref{alg2} in capturing certain image details. This contrast will be further validated in the experiments.
\begin{figure}[t]
    \centering
    \includegraphics[width=0.9\textwidth]{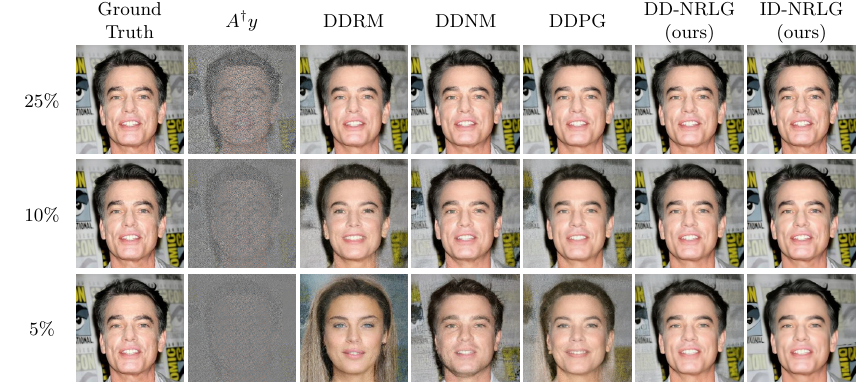} 
    \caption{\textbf{Qualitative results of noiseless compressive sensing} with different sampling rate (25\%,10\% and 5\%).}
  \label{fig3} 
\end{figure}

\section{Experiment}\label{sec4}

In this section, we validate the effectiveness of the proposed method for image restoration by designing several classic image linear inverse problems, including super-resolution, denoising, and deblurring (both Gaussian blur and motion blur).  Furthermore, our method achieves notably superior performance on compressed sensing tasks, with qualitative results shown in Figure \ref{fig3}. We compare our method with several state-of-the-art approaches, including DDRM \cite{ddrm}, DPS \cite{dps}, DDNM \cite{ddnm}, and DDPG \cite{ddpg}.

\subsection{Implementation Details}\label{sec4.1}

We conduct extensive experiments on two datasets: CelebA-HQ (256×256) and ImageNet (256×256). To ensure a fair comparison, all methods are evaluated in a zero-shot setting without any fine-tuning, using the same publicly available pre-trained models: the model trained on the CelebA-HQ dataset from \cite{lugmayr2022repaint}, and the unconditional model trained on ImageNet from \cite{imagenet}. Throughout all experiments, we adopt the same linear noise schedule $\left\{\beta_t\right\}$ for all methods. Each method is run for T=100 iterations, except for the DPS method, which requires T=1000 iterations. The degradation models are configured as follows:

\begin{itemize}
    \item For super-resolution, we apply bicubic downsampling with a scale factor of 4, as in \cite{ddrm};
    \item For Gaussian deblurring, a blur kernel is used with a standard deviation of 10 and a kernel size of 5×5;
    \item For denoising, we inject additive Gaussian noise with noise levels of $\left\{0.1, 0.25, 0.5\right\}$;
    \item For compressive sensing, we choose an orthogonalized random matrix  applied to the original image in a block-wise manner with the sampling ratios of $\left\{25\%, 10\%, 5\%\right\}$;
    \item For motion deblurring, we use the same motion blur kernel generated with a strength value of 0.5 as in \cite{ddpg}. For each observation model, except for denoising experiments, different levels of Gaussian noise are considered: $\left\{0, 0.05, 0.1\right\}$.
\end{itemize}

\textbf{Limitations of applicability:} Some methods have limitations for specific experiments. In the experimental results, "-" indicates that the method cannot solve the given problem. Specifically, DDRM is limited by separable kernels and SVD-based storage, making it unsuitable for motion deblurring. DDNM is designed for noise-free environments, and while its extended version, DDNM+ \cite{ddnm}, handles noisy inverse problems, it is still restricted by SVD storage and specific degradation models. Therefore, it is incompatible with noisy super-resolution and cannot effectively solve noisy deblurring problems (Appendix \ref{App.B.2}) in practice.
\begin{table}
    \centering
    \renewcommand{\arraystretch}{1.2} 
    \setlength{\tabcolsep}{4pt} 
    \caption{\textbf{Noiseless quantitative results} for several inverse problems on CelebA-HQ (top) and ImageNet (bottom). The best results are highlighted in \textbf{bold}, and the second-best results are \underline{underlined}.}
    \resizebox{\textwidth}{!}{ 
    \begin{tabular}{l ccc ccc ccc ccc}
        \toprule
        \textbf{CelebA-HQ} &\multicolumn{3}{c}{SR$\times$4} & \multicolumn{3}{c}{Deblurring(Gaussian)}&\multicolumn{3}{c}{CS(ratio=10\%)}&\multicolumn{3}{c}{CS(ratio=5\%)} \\
        \cmidrule(lr){2-4} \cmidrule(lr){5-7} \cmidrule(lr){8-10} \cmidrule(lr){11-13}
        \textbf{Method} & PSNR $\uparrow$ & LPIPS $\downarrow$ & SSIM $\uparrow$ & PSNR $\uparrow$ & LPIPS $\downarrow$ & SSIM $\uparrow$ & PSNR $\uparrow$ & LPIPS $\downarrow$ & SSIM $\uparrow$ & PSNR $\uparrow$ & LPIPS $\downarrow$ & SSIM $\uparrow$ \\
        \midrule
        DDRM\cite{ddrm}    & 31.98 & 0.057 & \underline{0.8848} & 45.57 & \underline{0.002} & 0.9885 & 25.59 & 0.160 & 0.7494 & 18.05 & 0.422 & 0.4836 \\ 
        DPS\cite{dps}      & 26.92 & 0.086 & 0.7467 & 30.34 & 0.049 & 0.8307 & - & - & - & - & - & - \\
        DDNM\cite{ddnm}    & 31.96 & \textbf{0.049} & 0.8811 & 49.10 & \textbf{0.001} & 0.9936 & 29.41 & 0.087 & 0.8319 & 23.25 & 0.239 & 0.6697 \\
        DDPG\cite{ddpg}    & 31.94 & \underline{0.051} & 0.8818 & 49.17 & \textbf{0.001} & 0.9942 & 28.85 & 0.123 & 0.8281 & 21.78 & 0.325 & 0.6371 \\
        \midrule
        DD-NRLG (ours)     & \underline{32.07} & \textbf{0.049} & 0.8834 & \underline{49.56} & \textbf{0.001} & \underline{0.9945} & \underline{31.67} & \underline{0.058} & \underline{0.8801} & \underline{29.03} & \underline{0.096} & \underline{0.8295} \\
        ID-NRLG (ours)     & \textbf{33.13} & 0.104 & \textbf{0.9101} & \textbf{49.90} & \textbf{0.001} & \textbf{0.9950} & \textbf{34.13} & \textbf{0.045} & \textbf{0.9236} & \textbf{30.98} & \textbf{0.090} & \textbf{0.8825} \\
        \bottomrule
        \\
        \toprule
        \textbf{ImageNet} &\multicolumn{3}{c}{SR$\times$4} & \multicolumn{3}{c}{Deblurring(Gaussian)}&\multicolumn{3}{c}{CS(ratio=10\%)}&\multicolumn{3}{c}{CS(ratio=5\%)} \\
        \cmidrule(lr){2-4} \cmidrule(lr){5-7} \cmidrule(lr){8-10} \cmidrule(lr){11-13}
        \textbf{Method} & PSNR $\uparrow$ & LPIPS $\downarrow$ & SSIM $\uparrow$ & PSNR $\uparrow$ & LPIPS $\downarrow$ & SSIM $\uparrow$ & PSNR $\uparrow$ & LPIPS $\downarrow$ & SSIM $\uparrow$ & PSNR $\uparrow$ & LPIPS $\downarrow$ & SSIM $\uparrow$ \\
        \midrule
        DDRM\cite{ddrm}    & 28.64 & \underline{0.188} & 0.8275 & 41.87 & \underline{0.004} & 0.9760 & 22.16 & 0.4382 & 0.5339 & 16.63 & 0.756 & 0.6025 \\ 
        DPS\cite{dps}      & 23.36 & 0.278 & 0.6405 & 28.07 & 0.167 & 0.7856 & - & - & - & - & - & - \\
        DDNM\cite{ddnm}    & 29.01 & 0.189 & 0.8355 & \underline{45.43} & \textbf{0.002} & 0.9884 & 25.04 & 0.299 & 0.6600 & 19.56 & 0.667 & 0.4154 \\
        DDPG\cite{ddpg}    & 29.00 & 0.196 & \textbf{0.8361} & 45.37 & \textbf{0.002} & \underline{0.9889} & 25.39 & 0.294 & 0.7054 & 19.50 & 0.644 & 0.4877 \\
        \midrule
        DD-NRLG (ours)     & \underline{29.05} & \textbf{0.187} & \textbf{0.8361} & 45.35 & \textbf{0.002} & 0.9886 & \underline{28.00} & \underline{0.160} & \underline{0.7829} & \underline{23.72} & \underline{0.355} & \underline{0.6529} \\
        ID-NRLG (ours)     & 28.80 & 0.258 & 0.8295 & \textbf{45.72} & \textbf{0.002} & \textbf{0.9890} & \textbf{31.11} & \textbf{0.105} & \textbf{0.8628} & \textbf{27.22} & \textbf{0.181} & \textbf{0.7827} \\
        \bottomrule
    \end{tabular}
    }
    \label{table2}
\end{table}
\subsection{Quantitative Results}\label{sec4.2}

For the quantitative experiments, we use several evaluation metrics, including Peak Signal-to-Noise Ratio (PSNR) to assess the fidelity of image restoration between two images, Learned Perceptual Image Patch Similarity (LPIPS) to represent the perceptual difference between two images, and Structural Similarity (SSIM) to measure the similarity in structural, luminance, and contrast information between the two images. As shown in Table \ref{table2} and Table\ref{table3}, our proposed methods achieve outstanding results across various image restoration problems on two datasets. 

\subsubsection{Results for noiseless problems }

In the noiseless experiments, we evaluate all methods on $4\times$ super-resolution, Gaussian deblurring, and compressive sensing at different sampling ratios across two datasets. The detailed results are presented in Table \ref{table2}. On the CelebA-HQ dataset, the DPS method consistently exhibits suboptimal performance in all tasks, while DDRM, DDNM, and DDPG achieve comparable results.

For super-resolution, our proposed ID-NRLG achieves the highest PSNR (33.13 dB) and SSIM (0.9101), outperforming all compared methods by a significant margin (1.1 dB in PSNR and 0.03 in SSIM). However, it slightly underperforms in the LPIPS metric, as the optimization scheme prioritizes fidelity and seeks the most optimal solution, with less emphasis on realism. In contrast, DD-NRLG achieves the best LPIPS score (0.049), on par with DDNM, while maintaining competitive PSNR performance. 

For Gaussian deblurring, both ID-NRLG and DD-NRLG demonstrate superior performance, with ID-NRLG achieving near-perfect LPIPS (0.001), matching DDNM, DDPG, and DD-NRLG, while also attaining the highest PSNR and SSIM, ensuring both fidelity and perceptual quality.

Similarly, on the ImageNet dataset, both proposed methods consistently perform well. In super-resolution and Gaussian deblurring, DD-NRLG and ID-NRLG achieve optimal results across all metrics. Other methods, such as DDNM and DDPG, deliver near-optimal performance, while DDRM shows a larger gap from the optimal results (SR$\times$4: PSNR ↓$\sim$0.4 dB, SSIM ↓$\sim$0.01; Gaussian Deblurring: PSNR ↓$\sim$4 dB, SSIM ↓$\sim$0.01). Notably, in super-resolution, ID-NRLG does not achieve the best fidelity as observed on the CelebA-HQ dataset. This is likely due to the higher complexity of the ImageNet dataset, which demands more accurate noise prediction. The absence of noise re-injection in ID-NRLG leads to higher prediction errors in the denoiser, which ultimately reduces result accuracy.

It is worth noting that our proposed methods exhibit excellent performance in noiseless compressive sensing tasks. Specifically, DD-NRLG and ID-NRLG consistently achieve the second-best and best results across all evaluation metrics, respectively. Even under extremely low sampling rate of 5\%, both methods are capable of producing high-quality reconstructions. This demonstrates the strong potential of our approach for practical applications, such as accelerated magnetic resonance imaging (MRI), low-dose computed tomography (CT), and other scenarios where efficient and reliable image reconstruction from limited measurements is crucial.

\begin{table}
    \centering
    \renewcommand{\arraystretch}{1.2} 
    \setlength{\tabcolsep}{4pt} 
    \caption{\textbf{Noisy quantitative results} for several inverse problems on CelebA-HQ (top) and ImageNet (bottom). The best results are highlighted in \textbf{bold}, and the second-best results are \underline{underlined}}
    \resizebox{\textwidth}{!}{ 
    \begin{tabular}{l ccc ccc ccc ccc ccc}
        \toprule
        \textbf{CelebA-HQ} &\multicolumn{3}{c}{SR$\times$4($\sigma_y=0.05$)} & \multicolumn{3}{c}{Deblurring(Gaussian,$\sigma_y=0.05$)}&\multicolumn{3}{c}{Denoising($\sigma_y=0.25$)}&\multicolumn{3}{c}{CS(ratio=10\%,$\sigma_y=0.05$)} &\multicolumn{3}{c}{Deblurring(Motion,$\sigma_y=0.05$)}\\
        \cmidrule(lr){2-4} \cmidrule(lr){5-7} \cmidrule(lr){8-10} \cmidrule(lr){11-13} \cmidrule(lr){14-16}
        \textbf{Method} & PSNR $\uparrow$ & LPIPS $\downarrow$ & SSIM $\uparrow$ & PSNR $\uparrow$ & LPIPS $\downarrow$ & SSIM $\uparrow$ & PSNR $\uparrow$ & LPIPS $\downarrow$ & SSIM $\uparrow$ & PSNR $\uparrow$ & LPIPS $\downarrow$ & SSIM $\uparrow$ & PSNR $\uparrow$ & LPIPS $\downarrow$ & SSIM $\uparrow$\\
        \midrule
        DDRM\cite{ddrm}    & 29.53 & \textbf{0.085} & 0.8313 & \underline{31.42} & 0.075 & \underline{0.8655} & 30.69 & \underline{0.083} & 0.8259 & 24.82 & 0.142 & 0.7474 & - & - & - \\ 
        DPS\cite{dps}      & 26.31 & \underline{0.097} & 0.7335 & 29.09 & 0.069 & 0.7985 & 28.59 & 0.084 & 0.8017 & - & - & - & 26.28 & 0.085 & 0.8017 \\
        DDNM\cite{ddnm}    & - & - & - & - & - & - & \underline{30.79} & \textbf{0.079} & \underline{0.8631} & - & - & - & - & - & - \\
        DDPG\cite{ddpg}    & \underline{29.61} & 0.102 & \underline{0.8321} & 30.81 & \underline{0.065} & 0.8496 & 28.43 & 0.168 & 0.8186 & 26.50 & 0.172 & 0.7923 & 29.45 & \underline{0.078} & \underline{0.8235}\\
        \midrule
        DD-NRLG (ours)     & 29.37 & \underline{0.097} & 0.8237 & 30.86 & \textbf{0.062} & 0.8440 & 30.43 & \textbf{0.079} & 0.8549 & \underline{30.28} & \textbf{0.077} & \underline{0.8485} & \underline{29.53} & \textbf{0.076} & 0.8206 \\
        ID-NRLG (ours)     & \textbf{30.09} & 0.169 & \textbf{0.8498} & \textbf{32.37} & 0.126 & \textbf{0.8827} & \textbf{31.04} & 0.103 & \textbf{0.8725} & \textbf{31.75} & \underline{0.079} & \textbf{0.8846} & \textbf{30.82} & 0.152 & \textbf{0.8588} \\
        \bottomrule
        \\
        \toprule
        \textbf{ImageNet} &\multicolumn{3}{c}{SR$\times$4($\sigma_y=0.05$)} & \multicolumn{3}{c}{Deblurring(Gaussian,$\sigma_y=0.05$)}&\multicolumn{3}{c}{Denoising($\sigma_y=0.25$)}&\multicolumn{3}{c}{CS(ratio=10\%,$\sigma_y=0.05$)}&\multicolumn{3}{c}{Deblurring(Motion,$\sigma_y=0.05$)} \\
        \cmidrule(lr){2-4} \cmidrule(lr){5-7} \cmidrule(lr){8-10} \cmidrule(lr){11-13} \cmidrule(lr){14-16}
        \textbf{Method} & PSNR $\uparrow$ & LPIPS $\downarrow$ & SSIM $\uparrow$ & PSNR $\uparrow$ & LPIPS $\downarrow$ & SSIM $\uparrow$ & PSNR $\uparrow$ & LPIPS $\downarrow$ & SSIM $\uparrow$ & PSNR $\uparrow$ & LPIPS $\downarrow$ & SSIM $\uparrow$ & PSNR $\uparrow$ & LPIPS $\downarrow$ & SSIM $\uparrow$ \\
        \midrule
        DDRM\cite{ddrm}    & 26.83 & \textbf{0.231} & \underline{0.7660} & 29.10 & \textbf{0.164} & 0.8227 & 28.90 & \textbf{0.135} & 0.8259 & 22.06 & 0.403 & 0.5655 & - & - & - \\ 
        DPS\cite{dps}      & 23.05 & \underline{0.284} & 0.6307 & 26.88 & \underline{0.185} & 0.7540 & 25.96 & 0.234 & 0.7357 & - & - & - & 23.28 & 0.254 & 0.6416\\
        DDNM\cite{ddnm}    & - & - & - & - & - & - & \underline{29.25} & 0.163 & \underline{0.8379} & - & - & - & - & - & - \\
        DDPG\cite{ddpg}    & \underline{26.84} & 0.291 & 0.7660 & \underline{29.14} & \textbf{0.164} & \textbf{0.8257} & 26.86 & 0.236 & 0.7296 & 23.63 & 0.364 & 0.6765 & \underline{27.33} & \textbf{0.190} & \underline{0.7727} \\
        \midrule
        DD-NRLG (ours)     & \textbf{26.87} & 0.306 & \textbf{0.7666} & \textbf{29.31} & 0.220 & \underline{0.8244} & 29.14 & \underline{0.142} & 0.8374 & \underline{27.13} & \underline{0.191} & \underline{0.7101} & \textbf{27.80} & \underline{0.252} & \textbf{0.7847}\\
        ID-NRLG (ours)     & 25.77 & 0.405 & 0.7213 & 28.38 & 0.306 & 0.8023 & \textbf{29.26} & 0.157 & \textbf{0.8394} & \textbf{29.28} & \textbf{0.138} & \textbf{0.8236} & 26.42 & 0.361 & 0.7432 \\
        \bottomrule
    \end{tabular}
    }
    \label{table3}
\end{table}
\subsubsection{Results for noisy problems}

In noisy settings with Gaussian noise, we evaluate the performance of methods on diverse linear inverse problems (super-resolution, Gaussian deblurring, denoising, compressive sensing and motion deblurring) across two datasets. Some experimental results are shown in Table \ref{table3} (complete results can be found in Appendix \ref{App.B}). When $\sigma_{\bm{y}} = 0.05$, on the CelebA-HQ dataset, DPS performs the worst. In super-resolution and Gaussian deblurring, DDRM and DDPG yield similar results. Our DD-NRLG slightly underperforms DDRM in the LPIPS for super-resolution (0.097 vs. 0.085), but achieves the best performance in Gaussian deblurring (0.062), with PSNR and SSIM comparable to DDRM and DDPG. ID-NRLG shows significant improvements in PSNR and SSIM, especially in the Gaussian deblurring issue, where PSNR is about 1 dB higher, benefiting from the removal of noise re-injection during the generation process. In motion deblurring, compared to DPS and DDPG, ID-NRLG improves PSNR and SSIM by approximately 1 dB and 0.03, respectively.

In denoising, when the noise level is low ($\sigma_{\bm{y}}=0.1$), ID-NRLG outperforms DDNM in all metrics, with PSNR increasing by 0.6 dB and SSIM improving by 0.01. DD-NRLG performs similarly to DDRM and DDPG, but when the noise level increases ($\sigma_{\bm{y}}=0.25$), DDPG's performance declines significantly, indicating its limited robustness to strong noise. DD-NRLG achieves the best LPIPS score (0.079), comparable to DDNM, while ID-NRLG achieves the best PSNR and SSIM.

On the ImageNet dataset, our method shows a slight decline in performance but remains competitive. In denoising ($\sigma_{\bm{y}}=0.1$), ID-NRLG achieves the best performance across all metrics, with PSNR about 0.7 dB higher than the second-best DDNM, and LPIPS (0.055) lower than DDRM by 0.015. At a higher noise level ($\sigma_{\bm{y}}=0.25$), ID-NRLG still maintains the highest PSNR and SSIM, while DD-NRLG achieves the second-best LPIPS, just after DDRM .

In the super-resolution and Gaussian deblurring, the performance of our method declines. At the $\sigma_{\bm{y}}=0.05$ noise level, in super-resolution, only DD-NRLG achieves the best PSNR (26.87) and SSIM (0.7666), but its LPIPS is lower than DDRM and DDPG. In both Gaussian deblurring and motion deblurring experiments, LPIPS does not reach the optimal value and is still lower than DDPG and DDRM. We speculate that this is due to the more diverse image content and complex scenes in the ImageNet dataset, making it harder for the model to accurately capture image details and maintain global consistency during generation or restoration, thereby affecting performance.

Similarly, in noisy scenarios, our method continues to demonstrate strong performance in compressive sensing tasks, achieving the best and second-best results across various metrics, which highlights its robust resilience to noise.
\begin{figure}[t]
    \centering
    \includegraphics[width=\textwidth]{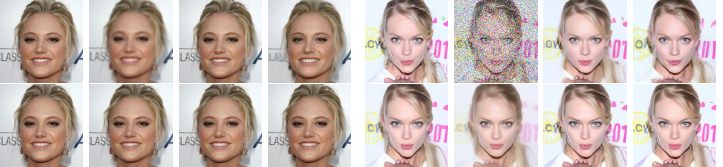} 
    \caption{\textbf{Qualitative results of super-resolition and denoising}. Left column: Super-resolution ×4 (noise level $\sigma_{\bm{y}} = 0$); Right column: Denoising (noise level $\sigma_{\bm{y}}=0.5$). Top row: original, measurement, DDRM, DPS. Bottom row: DDNM, DDPG, DD-NRLG (ours), ID-NRLG (ours). }
  \label{fig4} 
\end{figure}

\subsubsection{Summary}

Overall, DPS performs poorly across all problems, indicating a significant deviation in its likelihood score approximation. DDRM and DDPG exhibit stable performance but fail to demonstrate a distinct advantage in any specific task. DDNM excels in noise-free experiments but lacks generalization capability, struggling with noisy degradations beyond denoising. This may be due to the amplification of noise in the degraded measurement $A^{\dagger}\bm{y}$ making the algorithm less robust to noise.

In contrast, our proposed methods demonstrate stronger robustness to noise and offer clear advantages. The sampling-based method, DD-NRLG, prioritizes a balance between realism and consistency in image generation, producing images that align more closely with human visual perception and achieving superior performance in LPIPS. The iterative optimization method, ID-NRLG, focuses on identifying the optimal solution that best matches the observed data, excelling in fidelity metrics such as PSNR and SSIM.

\vspace{5pt}
\textit{\textbf{Remark:}} It is important to emphasize that our methods can achieve high-quality reconstructions even at extremely low sampling rates in compressive sensing tasks, demonstrating strong recovery capability and robustness. This characteristic gives the method with significant potential value in practical applications such as medical imaging, remote sensing, and communications. More experimental results can be found in the Appendix \ref{App.B}.

\subsection{Qualitative Results}\label{sec4.3}

\begin{figure}[t]
  \centering
  \includegraphics[width=\textwidth]{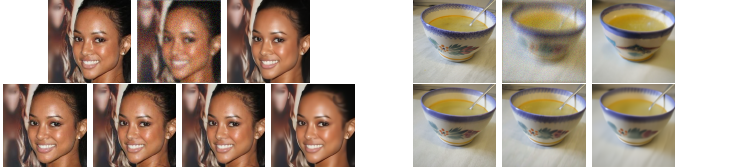} 
  \caption{\textbf{Qualitative results of deblurring}. Left column: Gaussian Deblurring ($\sigma_{\bm{y}} = 0.1$). Top row: original, measurement, DDRM. Bottom row: DPS, DDPG, DD-NRLG (ours), ID-NRLG (ours). Right column: Motion Deblurring ($\sigma_{\bm{y}} = 0.05$). Top row: original, measurement, DPS. Bottom row: DDPG, DD-NRLG (ours), ID-NRLG (ours). }
  \label{fig5}
\end{figure}

As shown in Figure \ref{fig3} - Figure \ref{fig5} and Appendix \ref{App.B.3}, our method achieves high-quality image restoration.
Figure \ref{fig3} shows the results of compressive sensing at different sampling rates under noiseless conditions. It can be seen that our method achieves the best performance across all sampling rates. In particular, at a 5\% sampling rate, while other methods such as DDRM, DDNM, and DDPG exhibit obvious artifacts and unrealistic reconstructions, our method still delivers high-quality recovery. 

Figure \ref{fig4} presents the results for super-resolution under noise-free conditions and denoising with noise level $\sigma_{\bm{y}}=0.5$. In super-resolution, the DPS method yields blurry backgrounds and suboptimal restoration quality, while DDRM, DDNM, DDPG, and DD-NRLG yield similar results. However, ID-NRLG excels in background restoration, achieving significantly clearer text recovery.

For the high-noise-level denoising experiments, DPS introduces more artifacts, while DDRM, DDNM, and DDPG fail to maintain consistency with the ground truth in restoring background letters and numbers. In contrast, our methods demonstrate superior restoration performance. However, it can also be observed that ID-NRLG, which prioritizes data fidelity, may compromise sample realism, leading to noticeable anomalies in areas such as the person's hand.

Figure \ref{fig5} presents the restoration results for Gaussian deblurring with a noise level of $\sigma_{\bm{y}}=0.1$ and motion deblurring with a noise level of $\sigma_{\bm{y}}=0.05$. For Gaussian deblurring, DPS, DDPG, and our DD-NRLG provide better realism, with higher-quality restoration of texture details like the hair, while DDRM and ID-NRLG are smoother. In the more complex motion deblurring problems, the results of DPS show a larger difference from the ground truth, while our method, compared to DDPG, recovers finer details, such as the pattern on the edge of the bowl, more effectively.

Similarly, through qualitative observation, we can also see that the DD-NRLG method focuses more on perceptual metrics, resulting in a restoration that appears more visually realistic, while ID-NRLG emphasizes data fidelity. 
\subsection{Ablation Study}\label{sec4.4}

\subsubsection{Effect of Removing Jacobian Matrix Computation}

As introduced in Section \ref{sec3.1}, to enhance computational efficiency, we assume that the output $\bm{\epsilon}_{\theta}(\bm{x}_t,t)$ of the noise prediction network is independent of the sampling state $\bm{x}_t$, thereby eliminating the need for backpropagation. Here, we experimentally examine the validity of this assumption.

In Table \ref{table4}, we compare our method with the version that involves Jacobian matrix computation in terms of both restoration performance and computational time under a noisy setting ($\sigma_y=0.05$) across two datasets. As shown, removing the Jacobian matrix computation leads to only minor decreases in PSNR and LPIPS across various models, while significantly reducing computational time. This validates the reasonableness of assuming independence between the noise prediction network’s output $\bm{\epsilon}_{\theta}(\bm{x}_t,t)$ and the sampling state $\bm{x}_t$

\begin{table}[h]
    \centering
    \renewcommand{\arraystretch}{1.2} 
    \setlength{\tabcolsep}{8pt} 
    \caption{Restoration and Time Comparison: With vs. Without Jacobian Computation}
    \resizebox{0.9\textwidth}{!}{ 
    \begin{tabular}{l c ccc ccc}
        \toprule
        &&\multicolumn{3}{c}{\textbf{SR$\times$4(0.05,CelebA-HQ)}} & \multicolumn{3}{c}{\textbf{SR$\times$4(0.05,ImageNet)}} \\
        \cmidrule(lr){3-5} \cmidrule(lr){6-8}
        \textbf{Method} & \textbf{Compute Jacobian?}& PSNR $\uparrow$ & LPIPS $\downarrow$ & Time & PSNR $\uparrow$ & LPIPS $\downarrow$ & Time \\
        \midrule
        DD-NRLG    & \checkmark & 29.48 & 0.100 & 32.26s & 26.91 & 0.304 & 72.94s \\
        DD-NRLG    & \ding{55} & 29.37 & 0.097 & 8.05s & 26.87 & 0.306 & 18.79s \\ 
        \midrule
        Comparison    & - & \textcolor{red}{\textbf{$\downarrow$0.11}} & \textcolor{blue}{\textbf{$\downarrow$0.003}} & \textcolor{blue}{\textbf{$\downarrow$24.21s}} & \textcolor{red}{\textbf{$\downarrow$0.04}} & \textcolor{red}{\textbf{$\uparrow$0.002}} & \textcolor{blue}{\textbf{$\downarrow$54.15s}} \\
        \bottomrule
    \end{tabular}
    }
    \label{table4}
\end{table}

\subsubsection{Effect of Mean Correction} 

In Section \ref{sec3.1}, we adjust the mean of the derived posterior distribution $p(\bm{x}_{0}|\bm{x}_{t})$ using the posterior mean obtained from Tweedie's formula (Eq.(\ref{eq15})). To evaluate the impact of this adjustment, we conduct an ablation study. Specifically, we perform experiments on the CelebA-HQ dataset for bicubic super-resolution and Gaussian deblurring, under both noisy and noise-free conditions. We compare the performance of our two proposed schemes with the corrected mean against the original posterior mean.

\begin{figure}[h]
    \centering
    \includegraphics[width=0.5\textwidth]{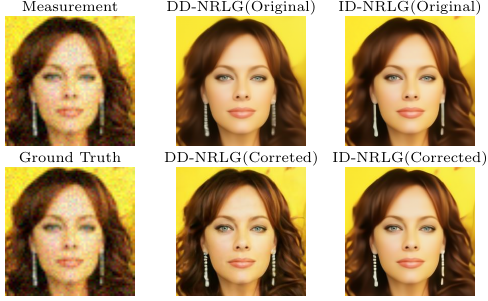} 
    \caption{ Super-Resolution with noise level 0.05. Ablation study about the effect of mean correction on restoration performance.}
    \label{fig6}
\end{figure}

As shown in Table \ref{table5} and Figure \ref{fig6}, the restoration using the corrected mean generally outperforms that using the original mean, particularly in noise-free settings, where the performance improvement is more significant. From a qualitative perspective, using the corrected mean results in better perceptual realism, with more detailed information being reconstructed, such as the earrings and hair texture of the person in the image, with higher restoration quality when the corrected mean is used.

\begin{table}[h]
    \centering
    \renewcommand{\arraystretch}{1.2} 
    \setlength{\tabcolsep}{8pt} 
    \caption{Ablation Study Results for Super-Resolution and Gaussian Deblurring on CelebA-HQ (PSNR(↑)/LPIPS(↓)). "Original" uses the original posterior mean, while "Corrected" uses the calibrated posterior mean.}
    \resizebox{0.65\textwidth}{!}{ 
    \begin{tabular}{l cc cc}
        \toprule
        & \multicolumn{2}{c}{\textbf{DD-NRLG}} & \multicolumn{2}{c}{\textbf{ID-NRLG}} \\
        \cmidrule(lr){2-3} \cmidrule(lr){4-5}
        & Original & Corrected & Original & Corrected \\
        \midrule
        SR$\times$4(0)    & 31.45/0.085 & \textbf{\underline{32.07}}/\textbf{\underline{0.049}} & 32.95/0.130 & \textbf{\underline{33.13}}/\textbf{\underline{0.104}} \\
        SR$\times$4(0.05) & 28.77/0.099 & \textbf{\underline{29.37}}/\textbf{\underline{0.097}} & 29.94/0.192 & \textbf{\underline{30.09}}/\textbf{\underline{0.169}} \\
        \midrule
        G.Deb(0)    & 42.28/0.009 & \textbf{\underline{49.56}}/\textbf{\underline{0.001}} & 42.36/0.008 & \textbf{\underline{49.90}}/\textbf{\underline{0.001}} \\
        G.Deb(0.05) & 30.70/0.067 & \textbf{\underline{30.86}}/\textbf{\underline{0.062}} & 31.94/0.161 & \textbf{\underline{32.37}}/\textbf{\underline{0.126}} \\
        \bottomrule
    \end{tabular}
    }
    \label{table5}
\end{table}
\subsubsection{The Role of Refining Noise}

In Section \ref{sec3.2}, the guiding mechanism we propose aims to refine the noise rather than directly adjust the sampled images. Based on this, we design the following ablation study to explore the impact of this approach: the guiding mechanism is applied to directly adjust the sampled images, and its performance is compared with that of our method. The experiments are conducted on the CelebA-HQ dataset, including super-resolution and denoising.

\begin{table}[h]
\centering 
\setlength{\tabcolsep}{8pt} 
\renewcommand{\arraystretch}{1} 
\caption{Ablation Study Results for Direct Image Adjustment vs. Noise Refinement(DD-NRLG) on the CelebA-HQ Dataset ( PSNR (↑) / LPIPS (↓) / SSIM(↑) ) for Super-Resolution and Denoising problems.}
\vspace{0.2em} 
\resizebox{0.5\textwidth}{!}{ 
\begin{tabular}{@{\extracolsep\fill}lcc@{\extracolsep\fill}}
\toprule
 & Image Adjustment & DD-NRLG \\ 
\midrule
SR$\times$4(0) & 30.55/0.061/0.8422 & \textbf{\underline{32.07}}/\textbf{\underline{0.049}}/\textbf{\underline{0.8834}}\\ 
SR$\times$4(0.05) & 28.05/\textbf{\underline{0.091}}/0.7718 & \textbf{\underline{29.37}}/0.097/\textbf{\underline{0.8273}} \\ 
\midrule
Deno(0.1) &32.82/0.054/0.8894 & \textbf{\underline{33.51}}/\textbf{\underline{0.046}}/\textbf{\underline{0.9037}}\\ 
Deno(0.25) & 29.35/0.086/0.8154 & \textbf{\underline{30.43}}/\textbf{\underline{0.079}}/\textbf{\underline{0.8549}}\\
\bottomrule
\end{tabular}
}
\label{table6}
\end{table} 
\begin{figure}[h]
    \centering
    \includegraphics[width=0.6\textwidth]{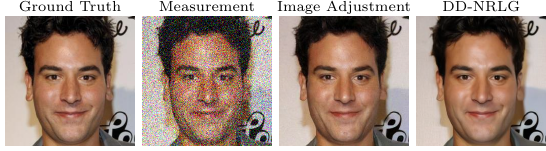} 
    \caption{ Denoising with noise level 0.25. Ablation study about the effect of refining noise on restoration performance.}
    \label{fig7}
\end{figure}

The corresponding experimental results are presented in Table \ref{table6} and Figure \ref{fig7}. The results show that the noise optimization-based approach not only performs better in quantitative metrics but also yields superior image restoration results, as seen in Figure \ref{fig7}. The restored images exhibit higher realism, with more detailed recovery of features such as facial texture, hair, and ears. Additionally, due to the use of the guiding mechanism for direct image adjustment, we employed a DDPM-based sampling scheme, which incurs higher computational cost.

\section{Conclusions}\label{sec5}

In this paper, we propose an innovative diffusion model guidance method that does not require task-specific training for different inverse problems. It directly utilizes a single pretrained model to solve various linear imaging inverse problems, demonstrating strong zero-shot generalization ability. Based on Tweedie's formula and forward diffusion process, we derive an approximate expression for the likelihood score and design an unconditional noise refinement method that incorporates conditional information into the noise, avoiding direct correction of sampled results. This makes image restoration more consistent with the diffusion model’s generative mechanism, improving reconstruction quality. We also propose two generation schemes based on sampling and iterative optimization, and validate the effectiveness of our method through simulation experiments. Notably, our method performs excellently in compressive sensing tasks, especially achieving high-quality reconstruction even at extremely low sampling rates, demonstrating strong recovery capability and noise robustness, with potential applications in medical imaging, remote sensing, and communications.

\textbf{Limitations\&Future Work:} Although our method achieves competitive restoration performance, it still has certain limitations. The derivation of the posterior distribution requires prior information about the data. In this paper, we only incorporate prior information into the posterior mean and do not correct for the variance. Furthermore, when deriving the approximate analytical expression for the likelihood score, for computational efficiency, we treat the noise predictor as independent of the variable $\bm{x}_{t}$ and ignore the derivative of the noise predictor, which inevitably introduces errors. Therefore, exploring methods to correct the variance and reduce the errors caused by ignoring the gradient of the noise predictor will be key areas for future research.

\section*{Acknowledgements}
This work was supported in part by the National Natural Science Foundation of China under Grant 61971223, in part by the Key Laboratory of Analysis of Mathematical Theory and Modeling of Complex Systems, the Ministry of Industry and Information Technology.

\bibliographystyle{unsrtnat}
\bibliography{references} 

\section*{Appendix}

\appendix
\counterwithin{figure}{section}  
\counterwithin{table}{section}  
\renewcommand{\thefigure}{\Alph{section}.\arabic{figure}}  
\renewcommand{\thetable}{\Alph{section}.\arabic{table}} 

\section{Supplementary proofs}\label{App.A}

\subsection{Optimality of Tweedie's Formula}\label{App.A.1}

Now, we will prove that the posterior mean obtained from Tweedie's formula is the optimal posterior mean. Firstly, we want to minimize the mean squared error between the estimate \( \hat{\bm{x}}_{0} \) and the ground truth \( \bm{x}_{0} \):
\[
\mathbb{E} \left[ \| \bm{x}_{0} - \hat{\bm{x}}_{0} \|^2 \right],
\]
where \( \hat{\bm{x}}_{0} \) is an estimate of \( \bm{x}_{0} \), and \( \bm{x}_{0} \) is the true noiseless signal.

Introducing the conditional expectation \( \mathbb{E}[\bm{x}_{0} | \bm{x}_{t}] \), we decompose the error as:

\[
\bm{x}_{0} - \hat{\bm{x}}_{0} = (\bm{x}_{0} - \mathbb{E}[\bm{x}_{0} | \bm{x}_{t}]) + (\mathbb{E}[\bm{x}_{0} | \bm{x}_{t}] - \hat{\bm{x}}_{0}).
\]

Taking the squared expectation:
\[
\mathbb{E} \left[ \| \bm{x}_{0} - \hat{\bm{x}}_{0} \|^2 \right] = \mathbb{E} \left[ \| (\bm{x}_{0} - \mathbb{E}[\bm{x}_{0} | \bm{x}_{t}]) + (\mathbb{E}[\bm{x}_{0} | \bm{x}_{t}] - \hat{\bm{x}}_{0}) \|^2 \right],
\]
\[
\mathbb{E} \left[ \| \bm{x}_{0} - \hat{\bm{x}}_{0} \|^2 \right] = \mathbb{E} \left[ \| \bm{x}_{0} - \mathbb{E}[\bm{x}_{0} | \bm{x}_{t}] \|^2 \right] + \mathbb{E} \left[ \| \mathbb{E}[\bm{x}_{0} | \bm{x}_{t}] - \hat{\bm{x}}_{0} \|^2 \right] + 2 \mathbb{E} \left[ (\bm{x}_{0} - \mathbb{E}[\bm{x}_{0} | \bm{x}_{t}])^\top (\mathbb{E}[\bm{x}_{0} | \bm{x}_{t}] - \hat{\bm{x}}_{0}) \right].
\]

Since \( \bm{x}_{0} - \mathbb{E}[\bm{x}_{0} | \bm{x}_{t}] \) has zero conditional expectation, the cross-term vanishes:
\[
\mathbb{E} \left[ \| \bm{x}_{0} - \hat{\bm{x}}_{0} \|^2 \right] = \mathbb{E} \left[ \| \bm{x}_{0} - \mathbb{E}[\bm{x}_{0} | \bm{x}_{t}] \|^2 \right] + \mathbb{E} \left[ \| \mathbb{E}[\bm{x}_{0} | \bm{x}_{t}] - \hat{\bm{x}}_{0} \|^2 \right].
\]

The first term $\mathbb{E} \left[ \| \bm{x}_{0} - \mathbb{E}[\bm{x}_{0} | \bm{x}_{t}] \|^2 \right]$ represents the unavoidable error caused by the insufficient information in \( \bm{x}_{t} \), which we cannot reduce. The second term $\mathbb{E} \left[ \| \mathbb{E}[\bm{x}_{0} | \bm{x}_{t}] - \hat{\bm{x}}_{0} \|^2 \right]$ represents the error between our estimated value \( \hat{\bm{x}}_{0} \) and the optimal estimate. 

Therefore, minimizing the mean squared error is equivalent to 
\[
\min_{\hat{\bm{x}}_{0}}\mathbb{E} \left[ \| \mathbb{E}[\bm{x}_{0} | \bm{x}_{t}] - \hat{\bm{x}}_{0} \|^2 \right].
\]
Clearly, the second term's error is minimized when $\hat{\bm{x}}_{0}=\mathbb{E}[\bm{x}_{0} | \bm{x}_{t}]$.
Therefore, the optimal posterior mean $\hat{\bm{x}}_{0}$ is the posterior mean $\mathbb{E}[\bm{x}_{0} | \bm{x}_{t}]$ obtained from Tweedie's formula.

\subsection{Proof of the property in Section \ref{sec3.1}}\label{APP.A.2}

Proving Eq.(\ref{eq23}) is equivalent to verifying the property of Gaussian marginals.

\vspace{10pt}
\textbf{Property:} If $ p(z_1 | z_0) = \mathcal{N}(z_0, V_1) $, $ p(z_2 | z_1) = \mathcal{N}(\alpha z_1, V_2) $, then $ p(z_2 | z_0) = \mathcal{N}(\alpha z_0, \alpha^2 V_1 + V_2) $.

\vspace{10pt}
\textit{Proof.} Firstly, the integral expression of the distribution$ p(z_2 | z_0) $ is:
\[
p(z_2 | z_0) = \int_{-\infty}^{\infty} 
  p(z_2 | z_1) \cdot p(z_1 | z_0) \, dz_1,
\]
Substituting the known Gaussian distribution, we obtain:

\[
p(z_2 | z_0) = \int_{-\infty}^{\infty} 
  \frac{1}{\sqrt{2\pi V_2}} \exp\left(-\frac{(z_2 - \alpha z_1)^2}{2V_2}\right) 
  \cdot \frac{1}{\sqrt{2\pi V_1}} \exp\left(-\frac{(z_1 - z_0)^2}{2V_1}\right) 
  \, dz_1.
\]
By combining the exponents, we obtain:

\begin{align}
\text{exponents} &= -\frac{(z_2 - \alpha z_1)^2}{2V_2} - \frac{(z_1 - z_0)^2}{2V_1} \notag\\
&= -\frac{z_2^2 - 2\alpha z_1 z_2 + \alpha^2 z_1^2}{2V_2} 
  - \frac{z_1^2 - 2 z_0 z_1 + z_0^2}{2V_1} \notag\\
&= -\frac{\alpha^2 z_1^2}{2V_2} - \frac{z_1^2}{2V_1} 
  + \frac{2\alpha z_1 z_2}{2V_2} - \frac{2 z_0 z_1}{2V_1} 
  - \frac{z_2^2}{2V_2} - \frac{z_0^2}{2V_1} \notag\\
&= -\frac{z_1^2\left(\frac{1}{V_1} + \frac{\alpha^2}{V_2}\right)}{2} 
  + \frac{z_1 \left(\frac{2\alpha z_2}{V_2} + \frac{2z_0}{V_1}\right)}{2} 
  - \frac{z_2^2}{2V_2} - \frac{z_0^2}{2V_1} \notag.
\end{align}

Let $ A = \frac{1}{V_1} + \frac{\alpha^2}{V_2} $,$ B = \frac{\alpha z_2}{V_2} + \frac{z_0}{V_1} $, then the exponent can be written as:
\[
-\frac{A z_1^2 - 2B z_1}{2} - \frac{z_2^2}{2V_2} - \frac{z_0^2}{2V_1},
\]
\begin{align}
\text{exponents} &= -\frac{A z_1^2 - 2B z_1}{2} - \frac{z_2^2}{2V_2} - \frac{z_0^2}{2V_1} \notag\\
&=-\frac{A}{2}\left(z_1 - \frac{B}{A}\right)^2 + \frac{B^2}{2A} - \frac{z_2^2}{2V_2} - \frac{z_0^2}{2V_1}.\notag
\end{align}
Substitute the exponent into the integral expression:

\[
p(z_2 | z_0) = \frac{1}{2\pi\sqrt{ V_1 V_2}} 
  \cdot\exp\left(-\frac{z_0^2}{2V_1} - \frac{z_2^2}{2V_2} +\frac{B^2}{2A}\right) 
  \int_{-\infty}^{\infty} 
  \exp\left(-\frac{A}{2}\left(z_1 - \frac{B}{A}\right)^2\right) 
  \, dz_1.
\]
First, we handle the integral term $\int_{-\infty}^{\infty} 
  \exp\left(-\frac{A}{2}\left(z_1 - \frac{B}{A}\right)^2\right) 
  \, dz_1$. Let $u=z_1 - \frac{B}{A}$ and $t =\frac{A}{2}u^2 $, we have

\begin{align}
    \int_{-\infty}^{\infty} 
  \exp\left(-\frac{A}{2}\left(z_1 - \frac{B}{A}\right)^2\right) 
  \, dz_1 &=\int_{-\infty}^{\infty} 
  \exp\left(-\frac{A}{2}u^2\right) 
  \, du \notag \\
  &=2\int_{0}^{\infty} 
  \exp\left(-\frac{A}{2}u^2\right) 
  \, du \notag \\
  &=2\int_{0}^{\infty} 
  \exp\left(-t\right)\cdot \frac{1}{2}\sqrt{\frac{2}{A}}t^{-\frac{1}{2}} 
  \, dt \notag \\
  &=\sqrt{\frac{2}{A}}\int_{0}^{\infty} 
  \exp\left(-t\right)\cdot t^{-\frac{1}{2}}\, dt. \notag
\end{align}
By the properties of the Gamma function $\Gamma(\frac{1}{2})=\int_{0}^{\infty} 
  \exp\left(-t\right)\cdot t^{-\frac{1}{2}}\, dt=\sqrt{\pi}$, we obtain:
\[
\int_{-\infty}^{\infty} 
  \exp\left(-\frac{A}{2}\left(z_1 - \frac{B}{A}\right)^2\right) 
  \, dz_1=\sqrt{\frac{2\pi}{A}}.
\]

And then, substitute $A = \frac{V_2 + \alpha^2 V_1}{V_1 V_2}$ into the equation, and finally calculate the coefficient term:
\[
\frac{1}{\sqrt{2\pi(\alpha^2 V_1 + V_2)}}.
\]

Then, we consider the exponential term $\exp\left(-\frac{z_0^2}{2V_1} - \frac{z_2^2}{2V_2} +\frac{B^2}{2A}\right)$, substitute $A = \frac{V_2 + \alpha^2 V_1}{V_1 V_2},B=\frac{\alpha V_{1}z_{2} + V_{2}z_{0}}{V_{1}V_{2}}$ into the equation, we obtain:
\begin{align}
    \exp\left(-\frac{z_0^2}{2V_1} - \frac{z_2^2}{2V_2} +\frac{B^2}{2A}\right)&=\exp\left(-\frac{(V_{1}z_{2}^{2} + V_{2}z_{0}^{2})(\alpha^2 V_1+V_{2})}{2V_{1}V_{2}(\alpha^{2}V_{1}+V_{2})}+\frac{(\alpha z_{2}V_{1} +  z_{0}V_{2})^{2}}{2V_{1}V_{2}(\alpha^{2}V_{1}+V_{2})}\right) ,\notag\\
    &=\exp\left(-\frac{(z_{2}-\alpha z_{0})^{2}}{2(\alpha^{2}V_{1}+V_{2})}\right). \notag 
\end{align}
Therefore, in the end, we obtain:
\begin{align}
    p(z_2 | z_0) = \frac{1}{\sqrt{2\pi (\alpha^2 V_1 + V_2)}} 
  \exp\left(-\frac{(z_2 - \alpha z_0)^2}{2(\alpha^2 V_1 + V_2)}\right)=\mathcal{N}(\alpha z_0, \alpha^2 V_1 + V_2). \notag
\end{align}

\section{More Experimental Details and Results}\label{App.B}

In this section we present more details on the experiments, along wit more quantitative and qualitative results that were not included in the main body of the paper.

\subsection{Hyperparameter setting}\label{App.B.1}

As mentioned in Section \ref{sec4.1}, in our experiments we do not modify the denoising diffusion model hyperparameter $\left\{\beta_{t}\right\}$. Specifically, $\left\{\beta_{t}\right\}$ follows a linear schedule from $\beta_{start}=0.0001$ to $\beta_{end}=0.02$ and $T=100$ in our experiments. For denoising, we inject additive Gaussian noise with noise levels of $\left\{0.1, 0.25, 0.5\right\}$. For each observation model (excluding denoising) different levels of Gaussian noise are considered: $\left\{0, 0.05, 0.1\right\}$.

As described in Section \ref{sec4.1}, we conduct experiments on CelebA-HQ and ImageNet (256×256). Super-resolution uses bicubic downsampling (×4), Gaussian deblurring applies a 5×5 kernel with a standard deviation of 10, and motion deblurring uses a kernel with a strength of 0.5. Compressed sensing is tested under both noisy and noise-free conditions with sampling rates of $\left\{25\%, 10\%, 5\%\right\}$.

Compared to other methods, our approach requires tuning very few parameters: only the noise refinement step size 
and the parameter that controls the input of random and deterministic noise. As a result, the hyperparameter tuning process is relatively simple. The parameter settings for DD-NRLG are shown in Table \ref{tableB.1}.

\begin{table}[t]
\centering 
\setlength{\tabcolsep}{15pt}
\renewcommand{\arraystretch}{1} 
\caption{The hyperparameters of DD-NRLG for different inverse problems on two datasets.}
\vspace{0.2em} 
\resizebox{\textwidth}{!}{ 
\begin{tabular}{@{\extracolsep\fill}lcc@{\extracolsep\fill}}
\toprule
 & CelebA-HQ & ImageNet \\ 
\midrule
SR$\times$4(0) & $\mu=1.6, \zeta=0.75$ & $\mu=1.7, \zeta=0.7$\\ 
SR$\times$4(0.05) & $\mu=1.25, \zeta=0.9$ & $\mu=1, \zeta=1$ \\ 
\midrule
G.Deb(0) &$\mu=0.95, \zeta=1$ & $\mu=0.95, \zeta=1$\\
G.Deb(0.05) &$\mu=1, \zeta=0.8$ &$\mu=1, \zeta=1$\\
G.Deb(0.1) &$\mu=1, \zeta=0.8$ & $\mu=1, \zeta=1$\\
\midrule
Deno(0.1) &$\mu=0.7, \zeta=1$ & $\mu=0.7, \zeta=1$\\ 
Deno(0.25) & $\mu=1, \zeta=1$ & $\mu=0.9, \zeta=1$\\
Deno(0.5) & $\mu=1.2, \zeta=1$& $-$\\
\bottomrule
\end{tabular}\hspace{2mm}
\begin{tabular}{@{\extracolsep\fill}lcc@{\extracolsep\fill}}
\toprule
 & CelebA-HQ & ImageNet \\ 
\midrule
M.Deb(0.05) & $\mu=1, \zeta=0.8$ & $\mu=1, \zeta=1$\\
M.Deb(0.1) & $\mu=1, \zeta=0.8$ & $-$\\
\midrule
CS(ratio=5\%,0.) &$\mu=3.5, \zeta=1$ & $\mu=2.25, \zeta=1$\\
CS(ratio=10\%,0) &$\mu=3, \zeta=1$ &$\mu=2, \zeta=1$\\
CS(ratio=25\%,0) &$\mu=2.25, \zeta=1$ & $\mu=1.75, \zeta=1$\\
\midrule
CS(ratio=5\%,0.05) &$\mu=3.5, \zeta=1$ & $\mu=3.5, \zeta=1$\\ 
CS(ratio=10\%,0.05) & $\mu=2.2, \zeta=1$ & $\mu=2.75, \zeta=1$\\
CS(ratio=25\%,0.05) & $\mu=1.35, \zeta=1$& $\mu=1.75, \zeta=1$\\
\bottomrule
\end{tabular}
}
\label{tableB.1}
\end{table}

\subsection{More quantitative results}\label{App.B.2}
In Section \ref{sec4.2}, we only present partial experimental results in Tables \ref{table2} and \ref{table3}. In the following, we provide the complete experimental result table, which includes experiments under more noise levels. The full results can be found in Table \ref{tableB.2}. Our proposed sampling-based approach and iterative optimization-based approach provide more flexible choices for the objectives. DD-NRLG performs better in perceptual metrics (LPIPS), while ID-NRLG provides better results in PSNR/SSIM. In addition, we also conduct experiments to evaluate the applicability of DDNM\cite{ddnm}.

The complete experimental results demonstrate that our method performs well in both noise-free and high-noise settings, achieving competitive or even superior results compared to other methods. 

\subsubsection*{Failure of DDNM+.} 
\begin{figure}[h]
    \centering
    \includegraphics[width=0.75\textwidth]{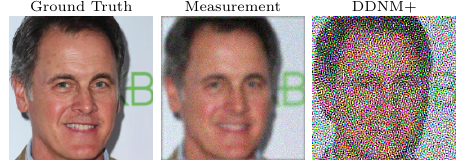} 
    \caption{ Failure of DDNM+ for Gaussian deblurring with noise level 0.05}
    \label{figB.1}
\end{figure}

As described in Section \ref{sec4.2}, DDNM was initially proposed to solve noiseless inverse problems, while its extended version, DDNM+, was designed to handle inverse problems with noise. However, DDNM+ processes noisy images $\bm{y}$ using Singular Value Decomposition (SVD), which is closely tied to specific downsampling cases and does not support situations like bicubic super-resolution and motion blur. When running noisy bicubic super-resolution or motion blur experimrnts, the official DDNM+ code outputs an "unsupported" assertion. Additionally, when handling noisy Gaussian deblurring problems, it also produces the failure results as shown in the Figure \ref{figB.1}.

DDNM+ can effectively handle denoising problems, but it fails in restoring noisy Gaussian deblurring. We believe this is because in the degradation modeling of the Gaussian deblurring, the degradation operator $\bm{A}$ is an irreversible linear operator. During the range null-space decomposition, the pseudoinverse operator $\bm{A}^{\dagger}$ in $\bm{A}^{\dagger}\bm{y}$ amplifies the noise, which makes the method not robust to noise. In contrast, the linear operator in the denoising problems is the identity matrix, so it does not amplify the noise in the degraded image. Therefore, DDNM+ can effectively handle denoising problems but cannot solve noisy Gaussian deblurring problems.

\subsubsection*{Comparison of Time Consumption}

Compared to existing zero-shot image restoration methods, our DD-NRLG also demonstrates significant advantages in terms of time consumption. We perform comparisons on the CelebA-HQ dataset for both 4× super-resolution and denoising, as shown in Table \ref{tableB.4}. In terms of efficiency for processing a single image, our method and DDPG are the fastest, with no noticeable change in computational efficiency. On the other hand, the DPS algorithm is the least efficient due to the expensive gradient computations and longer sampling steps (1000 NFEs). In contrast, our method achieves efficient restoration by analytically deriving the expression for the likelihood score and avoiding gradient calculations, combined with the DDIM sampling scheme.

\begin{table}[h]
\centering 
\setlength{\tabcolsep}{3pt} 
\renewcommand{\arraystretch}{2} 
\captionsetup{font=small} 
\caption{Computational Efficiency Comparison on CelebA-HQ. Experimental setup: Bicubic downsampling and additive Gaussian noise ($\sigma_{\bm{y}}=0.25$) with a batch\_size=1 on identical GPU. All methods use 100 sampling steps except DPS (1000 steps).}
\vspace{0.5em} 
\begin{tabular}{@{\extracolsep\fill}lccccc@{\extracolsep\fill}}
\toprule
 & DDRM & DPS & DDNM & DDPG & DD-NRLG(ours) \\ 
\midrule
SR$\times4$(0) & 8.13s & 230.23s & 8.15s & 8.07s & \textbf{\underline{8.02s}}\\ 
Deno(0.25) & 8.13s & 232.11s & 8.12s & 8.12s & \textbf{\underline{7.92s}}\\ 
\bottomrule
\end{tabular}
\label{tableB.4}
\end{table}
\hspace{10pt}

\begin{table}[h]
    \centering
    \renewcommand{\arraystretch}{1.2} 
    \setlength{\tabcolsep}{4pt} 
    \caption{All the experimental results}
    \resizebox{\textwidth}{!}{ 
    \begin{tabular}{l ccc ccc ccc ccc ccc}
        \toprule
        \textbf{CelebA-HQ} &\multicolumn{3}{c}{SR$\times$4($\sigma_y=0$)} & \multicolumn{3}{c}{SR$\times$4($\sigma_y=0.05$)}&\multicolumn{3}{c}{Deblurring(Gaussian,$\sigma_y=0$)}&\multicolumn{3}{c}{Deblurring(Gaussian,$\sigma_y=0.05$)} &\multicolumn{3}{c}{Deblurring(Gaussian,$\sigma_y=0.1$)}\\
        \cmidrule(lr){2-4} \cmidrule(lr){5-7} \cmidrule(lr){8-10} \cmidrule(lr){11-13} \cmidrule(lr){14-16}
        \textbf{Method} & PSNR $\uparrow$ & LPIPS $\downarrow$ & SSIM $\uparrow$ & PSNR $\uparrow$ & LPIPS $\downarrow$ & SSIM $\uparrow$ & PSNR $\uparrow$ & LPIPS $\downarrow$ & SSIM $\uparrow$ & PSNR $\uparrow$ & LPIPS $\downarrow$ & SSIM $\uparrow$ & PSNR $\uparrow$ & LPIPS $\downarrow$ & SSIM $\uparrow$\\
        \midrule
        DDRM\cite{ddrm}    & 31.98 & 0.057 & \underline{0.8848} & 29.53 & \textbf{0.085} & 0.8313 & 45.57 & \underline{0.002} & 0.9885 & \underline{31.42} & 0.075 & \underline{0.8655} & \underline{29.89} & 0.085 & \underline{0.8377} \\ 
        DPS\cite{dps}      & 26.92 & 0.086 & 0.7467 & 26.31 & \underline{0.097} & 0.7335 & 30.34 & 0.049 & 0.8307 & 29.09 & 0.069 & 0.7985 & 28.19 & 0.087 & 0.7760 \\
        DDNM\cite{ddnm}    & 31.96 & \textbf{0.049} & 0.8811 & - & - & - & 49.10 & \textbf{0.001} & 0.9936 & - & - & - & - & - & - \\
        DDPG\cite{ddpg}    & 31.94 & \underline{0.051} & 0.8818 & \underline{29.61} & 0.102 & \underline{0.8321} & 49.17 & \textbf{0.001} & 0.9942 & 30.81 & \underline{0.065} & 0.8496 & 29.46 & \textbf{0.076} & 0.8132\\
        \midrule
        DD-NRLG (ours)     & \underline{32.07} & \textbf{0.049} & 0.8834 & 29.37 & \underline{0.097} & 0.8237 & \underline{49.56} & \textbf{0.001} & \underline{0.9945} & 30.86 & \textbf{0.062} & 0.8440 & 29.49 & \underline{0.081} & 0.8193 \\
        ID-NRLG (ours)     & \textbf{33.13} & 0.104 & \textbf{0.9101} & \textbf{30.09} & 0.169 & \textbf{0.8498} & \textbf{49.90} & \textbf{0.001} & \textbf{0.9950} & \textbf{32.37} & 0.126 & \textbf{0.8827} & \textbf{30.56} & 0.168 & \textbf{0.8527} \\
        \bottomrule
        
        \\
        
        \toprule
        \textbf{CelebA-HQ} &\multicolumn{3}{c}{Denoising($\sigma_y=0.1$)} & \multicolumn{3}{c}{Denoising($\sigma_y=0.25$)}&\multicolumn{3}{c}{Denoising($\sigma_y=0.5$)}&\multicolumn{3}{c}{Deblurring(Motion,$\sigma_y=0.05$)}&\multicolumn{3}{c}{Deblurring(Motion,$\sigma_y=0.1$)} \\
        \cmidrule(lr){2-4} \cmidrule(lr){5-7} \cmidrule(lr){8-10} \cmidrule(lr){11-13} \cmidrule(lr){14-16}
        \textbf{Method} & PSNR $\uparrow$ & LPIPS $\downarrow$ & SSIM $\uparrow$ & PSNR $\uparrow$ & LPIPS $\downarrow$ & SSIM $\uparrow$ & PSNR $\uparrow$ & LPIPS $\downarrow$ & SSIM $\uparrow$ & PSNR $\uparrow$ & LPIPS $\downarrow$ & SSIM $\uparrow$ & PSNR $\uparrow$ & LPIPS $\downarrow$ & SSIM $\uparrow$ \\
        \midrule
        DDRM\cite{ddrm}    & 33.57 & 0.054 & 0.9090 & 30.69 & 0.083 & 0.8613 & 28.01 & 0.124 & 0.8135 & - & - & - & - & - & - \\ 
        DPS\cite{dps}      & 31.03 & 0.059 & 0.8857 & 28.59 & 0.084 & 0.8017 & 26.62 & \textbf{0.092} & 0.7545 & 26.28 & 0.085 & 0.7318 & 25.37 & \underline{0.097} & 0.7099\\
        DDNM\cite{ddnm}    & \underline{34.01} & 0.053 & \underline{0.9147} & \underline{30.79} & \textbf{0.079} & \underline{0.8631} & 28.19 & \underline{0.099} & 0.8147 & - & - & - & - & - & - \\
        DDPG\cite{ddpg}    & 33.52 & 0.055 & 0.8800 & 28.43 & 0.168 & 0.8186 & 25.47 & 0.179 & 0.7651 & 29.45 & \underline{0.078} & 0.8235 & 28.08 & \textbf{0.094} & 0.7886 \\
        \midrule
        DD-NRLG (ours)     & 33.51 & \underline{0.046} & 0.9037 & 30.43 & \textbf{0.079} & 0.8549 & \underline{28.20} & 0.121 & \underline{0.8162} & \underline{29.53} & \textbf{0.076} & 0.8206 & \textbf{28.13} & \underline{0.097} & \textbf{0.7961}\\
        ID-NRLG (ours)     & \textbf{34.61} & \textbf{0.039} & \textbf{0.9238} & \textbf{31.04} & 0.103 & \textbf{0.8725} & \textbf{28.22} & 0.162 & \textbf{0.8274} & \textbf{30.82} & 0.152 & \textbf{0.8588} & \textbf{28.94} & 0.188 & \textbf{0.8273} \\
        \bottomrule
    \end{tabular}
    }
    \label{tableB.2}
\end{table}
\begin{table}[h!]
    \centering
    \renewcommand{\arraystretch}{1.2} 
    \setlength{\tabcolsep}{4pt} 
    \resizebox{\textwidth}{!}{ 
    \begin{tabular}{l ccc ccc ccc ccc}
        \toprule
        \textbf{ImageNet} &\multicolumn{3}{c}{SR$\times$4($\sigma_y=0$)} & \multicolumn{3}{c}{SR$\times$4($\sigma_y=0.05$)}&\multicolumn{3}{c}{Denoising($\sigma_y=0.1$)}&\multicolumn{3}{c}{Denoising($\sigma_y=0.25$)} \\
        \cmidrule(lr){2-4} \cmidrule(lr){5-7} \cmidrule(lr){8-10} \cmidrule(lr){11-13}
        \textbf{Method} & PSNR $\uparrow$ & LPIPS $\downarrow$ & SSIM $\uparrow$ & PSNR $\uparrow$ & LPIPS $\downarrow$ & SSIM $\uparrow$ & PSNR $\uparrow$ & LPIPS $\downarrow$ & SSIM $\uparrow$ & PSNR $\uparrow$ & LPIPS $\downarrow$ & SSIM $\uparrow$ \\
        \midrule
        DDRM\cite{ddrm}    & 28.64 & \underline{0.188} & 0.8275 & 26.83 & \textbf{0.231} & \underline{0.7660} & 32.30 & \underline{0.070} & 0.8951 & 28.90 & \underline{0.135} & 0.8259 \\ 
        DPS\cite{dps}      & 23.36 & 0.278 & 0.6405 & 23.05 & \underline{0.284} & 0.6307 & 28.33 & 0.190 & 0.7917 & 25.96 & 0.234 & 0.7357 \\
        DDNM\cite{ddnm}    & \underline{29.01} & 0.189 & \underline{0.8355} & - & - & - & \underline{33.18} & 0.076 & \underline{0.9099} & \underline{29.25} & 0.163 & \underline{0.8379} \\
        DDPG\cite{ddpg}    & 29.00 & 0.196 & \textbf{0.8361} & \underline{26.84} & 0.291 & \underline{0.7660} & 31.80 & 0.099 & 0.8465 & 26.86 & 0.236 & 0.7296 \\
        \midrule
        DD-NRLG (ours)     & \textbf{29.05} & \textbf{0.187} & \textbf{0.8361} & \textbf{26.87} & 0.306 & \textbf{0.7666} & 31.94 & 0.090 & 0.8724 & 29.14 & \underline{0.142} & 0.8374 \\
        ID-NRLG (ours)     & 28.80 & 0.258 & 0.8295 & 25.77 & 0.405 & 0.7213 & \textbf{33.83} & \textbf{0.055} & \textbf{0.9212} & \textbf{29.26} & 0.157 & \textbf{0.8394} \\
        \bottomrule
        
        \\
    
        \toprule
        \textbf{ImageNet} &\multicolumn{3}{c}{Deblurring(Gaussian,$\sigma_y=0$)} & \multicolumn{3}{c}{Deblurring(Gaussian,$\sigma_y=0.05$)}&\multicolumn{3}{c}{Deblurring(Gaussian,$\sigma_y=0.1$)}&\multicolumn{3}{c}{Deblurring(Motion,$\sigma_y=0.05$)} \\
        \cmidrule(lr){2-4} \cmidrule(lr){5-7} \cmidrule(lr){8-10} \cmidrule(lr){11-13}
        \textbf{Method} & PSNR $\uparrow$ & LPIPS $\downarrow$ & SSIM $\uparrow$ & PSNR $\uparrow$ & LPIPS $\downarrow$ & SSIM $\uparrow$ & PSNR $\uparrow$ & LPIPS $\downarrow$ & SSIM $\uparrow$ & PSNR $\uparrow$ & LPIPS $\downarrow$ & SSIM $\uparrow$ \\
        \midrule
        DDRM\cite{ddrm}    & 41.87 & \underline{0.004} & 0.9760 & 39.10 & \textbf{0.164} & 0.8227 & \underline{27.38} & \underline{0.215} & \underline{0.7773} & - & - & - \\ 
        DPS\cite{dps}      & 28.07 & 0.167 & 0.7856 & 26.88 & \underline{0.185} & 0.7540 & 25.88 & \textbf{0.202} & 0.7239 & 23.28 & 0.254 & 0.6416 \\
        DDNM\cite{ddnm}    & \underline{45.43} & \textbf{0.002} & 0.9884 & - & - & - & - & - & - & - & - & - \\
        DDPG\cite{ddpg}    & 45.37 & \textbf{0.002} & \underline{0.9889} & \underline{29.14} & \textbf{0.164} & \underline{0.8257} & 27.30 & 0.308 & 0.7677 & \underline{27.33} & \textbf{0.190} & \underline{0.7727} \\
        \midrule
        DD-NRLG (ours)     & 45.35 & \textbf{0.002} & 0.9886 & \textbf{29.31} & 0.220 & \underline{0.8224} & \textbf{27.54} & 0.292 & \textbf{0.7776} & \textbf{27.80} & \underline{0.252} & \textbf{0.7847} \\
        ID-NRLG (ours)     & \textbf{45.72} & \textbf{0.002} & \textbf{0.9890} & 28.38 & 0.306 & 0.8023 & 26.07 & 0.407 & 0.7316 & 26.42 & 0.361 & 0.7432 \\
        \bottomrule
    \end{tabular}
    }
\end{table}
\begin{table}[h]
    \centering
    \renewcommand{\arraystretch}{1.2} 
    \setlength{\tabcolsep}{4pt} 
    \resizebox{\textwidth}{!}{ 
    \begin{tabular}{l ccc ccc ccc ccc ccc}
        \toprule
        \textbf{CelebA-HQ} &\multicolumn{3}{c}{CS(ratio=5\%,$\sigma_y=0$)} & \multicolumn{3}{c}{CS(ratio=10\%,$\sigma_y=0$)}&\multicolumn{3}{c}{CS(ratio=25\%,$\sigma_y=0$)}&\multicolumn{3}{c}{CS(ratio=5\%,$\sigma_y=0.05$)} &\multicolumn{3}{c}{CS(ratio=25\%,$\sigma_y=0.05$)}\\
        \cmidrule(lr){2-4} \cmidrule(lr){5-7} \cmidrule(lr){8-10} \cmidrule(lr){11-13} \cmidrule(lr){14-16}
        \textbf{Method} & PSNR $\uparrow$ & LPIPS $\downarrow$ & SSIM $\uparrow$ & PSNR $\uparrow$ & LPIPS $\downarrow$ & SSIM $\uparrow$ & PSNR $\uparrow$ & LPIPS $\downarrow$ & SSIM $\uparrow$ & PSNR $\uparrow$ & LPIPS $\downarrow$ & SSIM $\uparrow$ & PSNR $\uparrow$ & LPIPS $\downarrow$ & SSIM $\uparrow$\\
        \midrule
        DDRM\cite{ddrm}    & 18.05 & 0.422 & 0.4863 & 25.59 & 0.160 & 0.7494 & 33.10 & 0.0383 & 0.9.51 & 17.80 & 0.353 & 0.5267 & 31.12 & 0.057 & 0.8697 \\ 
        DDNM\cite{ddnm}    & 23.25 & 0.239 & 0.6697 & 29.41 & 0.087 & 0.8319 & 34.53 & 0.031 & 0.9210 & - & - & - & - & - & - \\
        DDPG\cite{ddpg}    & 21.78 & 0.325 & 0.6371 & 28.85 & 0.123 & 0.8281 & 34.82 & 0.038 & 0.9254 & 21.03 & 0.350 & 0.6043 & 31.26 & 0.073 & 0.8720\\
        \midrule
        DD-NRLG (ours)     & \underline{29.03} & \underline{0.096} & \underline{0.8295} & \underline{31.67} & \underline{0.058} & \underline{0.8801} & \underline{36.39} & \underline{0.020} & \underline{0.9437} & \underline{28.36} & \textbf{0.099} & \underline{0.8115} & \underline{32.57} & \underline{0.053} & \underline{0.8886} \\
        ID-NRLG (ours)     & \textbf{30.98} & \textbf{0.090} & \textbf{0.8825} & \textbf{34.13} & \textbf{0.045} & \textbf{0.9236} & \textbf{40.61} & \textbf{0.009} & \textbf{0.9751} & \textbf{29.99} & \underline{0.108} & \textbf{0.8560} & \textbf{33.69} & \textbf{0.039} & \textbf{0.9040} \\
        \bottomrule
        
        \\
        
        \toprule
        \textbf{ImageNet} &\multicolumn{3}{c}{CS(ratio=5\%,$\sigma_y=0$)} & \multicolumn{3}{c}{CS(ratio=10\%,$\sigma_y=0$)}&\multicolumn{3}{c}{CS(ratio=25\%,$\sigma_y=0$)}&\multicolumn{3}{c}{CS(ratio=5\%,$\sigma_y=0.05$)} &\multicolumn{3}{c}{CS(ratio=25\%,$\sigma_y=0.05$)} \\
        \cmidrule(lr){2-4} \cmidrule(lr){5-7} \cmidrule(lr){8-10} \cmidrule(lr){11-13} \cmidrule(lr){14-16}
        \textbf{Method} & PSNR $\uparrow$ & LPIPS $\downarrow$ & SSIM $\uparrow$ & PSNR $\uparrow$ & LPIPS $\downarrow$ & SSIM $\uparrow$ & PSNR $\uparrow$ & LPIPS $\downarrow$ & SSIM $\uparrow$ & PSNR $\uparrow$ & LPIPS $\downarrow$ & SSIM $\uparrow$ & PSNR $\uparrow$ & LPIPS $\downarrow$ & SSIM $\uparrow$ \\
        \midrule
        DDRM\cite{ddrm}    & 16.63 & 0.756 & 0.3025 & 22.16 & 0.438 & 0.5339 & 29.31 & 0.101 & 0.8070 & 16.47 & 0.739 & 0.3251 & 28.49 & 0.110 & 0.8017 \\ 
        DDNM\cite{ddnm}    & 19.56 & 0.667 & 0.4154 & 25.04 & 0.299 & 0.6600 & 30.50 & 0.083 & 0.8327 & - & - & - & - & - & - \\
        DDPG\cite{ddpg}    & 19.50 & 0.644 & 0.4877 & 25.39 & 0.294 & 0.7054 & 31.74 & 0.074 & 0.8725 & 18.88 & 0.7095 & 0.4319 & 29.15 & 0.167 & 0.8243 \\
        \midrule
        DD-NRLG (ours)     & \underline{23.72} & \underline{0.355} & \underline{0.6529} & \underline{28.00} & \underline{0.160} & \underline{0.7829} & \underline{32.69} & \underline{0.052} & \underline{0.8860} & \underline{24.49} & \underline{0.313} & \underline{0.6506} & \underline{29.29} & \underline{0.120} & \underline{0.7550}\\
        ID-NRLG (ours)     & \textbf{27.22} & \textbf{0.181} & \textbf{0.7827} & \textbf{31.11} & \textbf{0.105} & \textbf{0.8628} & \textbf{35.27} & \textbf{0.037} & \textbf{0.8628} & \textbf{26.78} & \textbf{0.217} & \textbf{0.7730} & \textbf{30.86} & \textbf{0.072} & \textbf{0.8158} \\
        \bottomrule
    \end{tabular}
    }
\end{table}

\subsection{More qualitative results}\label{App.B.3}
\begin{figure}[h]
    \centering
    \includegraphics[width=\textwidth]{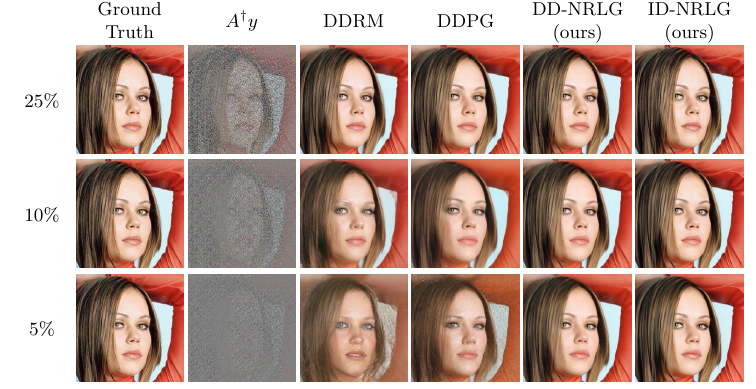} 
    \caption{CelebA-HQ: Compressive Sensing with noise level 0.05. }
    \label{figB.2}
\end{figure}

\begin{figure}[h]
    \centering
    \includegraphics[width=\textwidth]{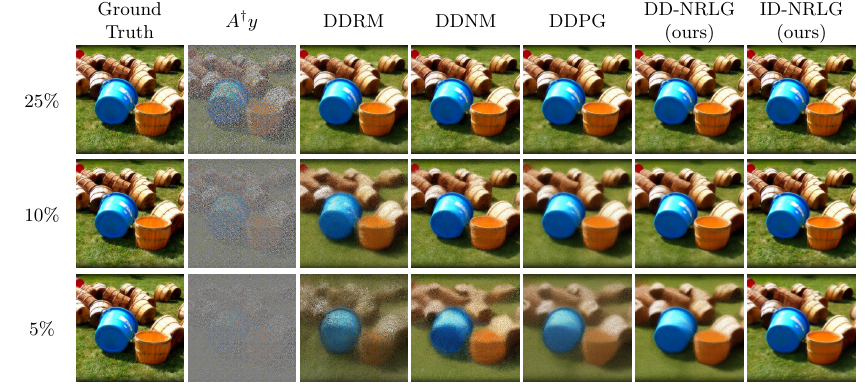} 
    \caption{ImageNet: Compressive Sensing with noise level 0. }
    \label{figB.3}
\end{figure}
\begin{figure}[h]
    \centering
    \includegraphics[width=\textwidth]{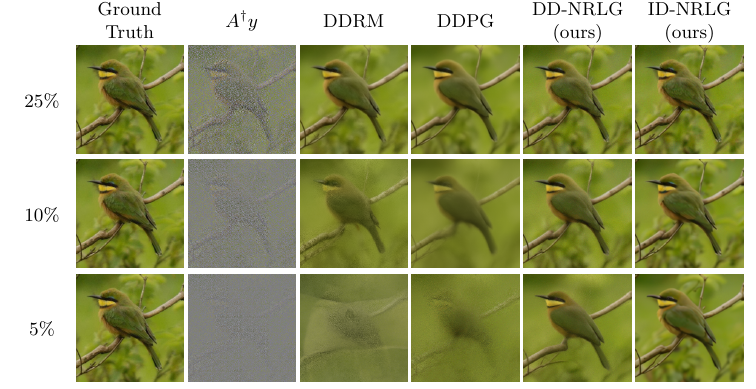} 
    \caption{ImageNet: Compressive Sensing with noise level 0.05. }
    \label{figB.4}
\end{figure}

\begin{figure}[h]
    \centering
    \includegraphics[width=0.75\textwidth]{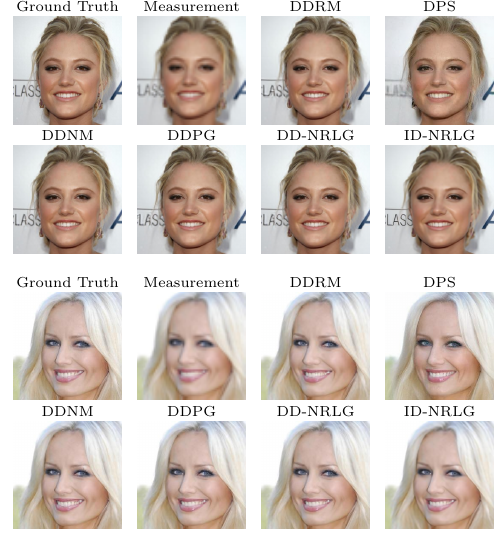} 
    \caption{CelebA-HQ: SR$\times$4 with noise level 0. }
    \label{figB.5}
\end{figure}

\begin{figure}[h]
    \centering
    \includegraphics[width=0.75\textwidth]{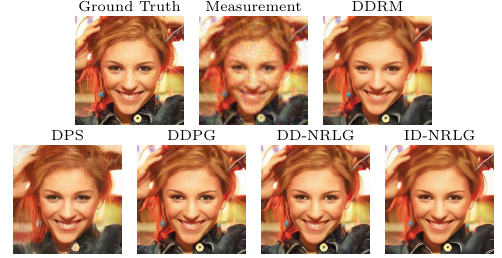} 
    \caption{CelebA-HQ: SR$\times$4 with noise level 0.05. }
    \label{figB.6}
\end{figure}

\begin{figure}[h]
    \centering
    \includegraphics[width=0.75\textwidth]{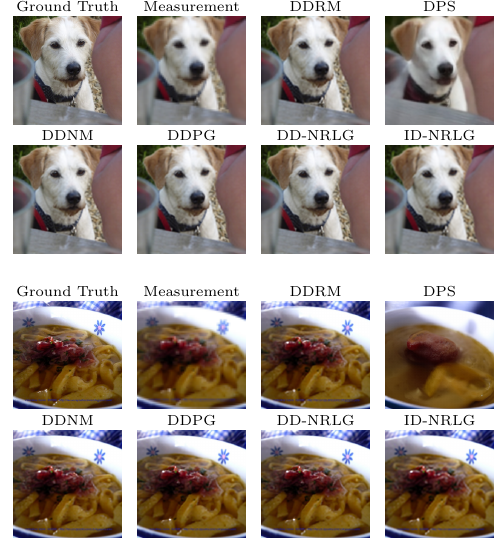} 
    \caption{ ImageNet: SR$\times$4 with noise level 0. }
    \label{figB.7}
\end{figure}

\begin{figure}[t]
    \centering
    \includegraphics[width=0.75\textwidth]{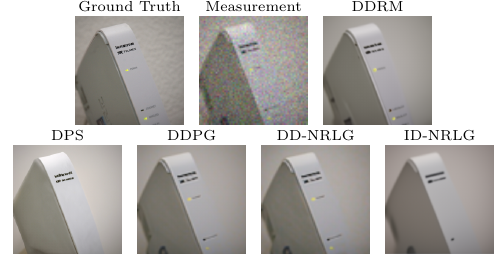} 
    \caption{ ImageNet: SR$\times$4 with noise level 0.05. }
    \label{figB.8}
\end{figure}

\begin{figure}[h]
    \centering
    \includegraphics[width=0.75\textwidth]{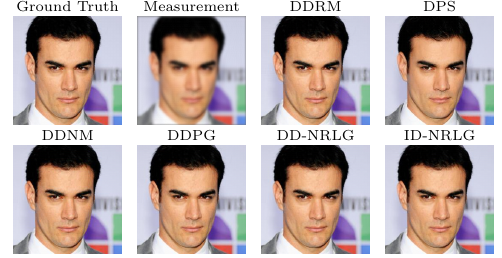} 
    \caption{ CelebA-HQ: Gaussiam Deblurring with noise level 0. }
    \label{figB.9}
\end{figure}

\begin{figure}[h]
    \centering
    \includegraphics[width=0.75\textwidth]{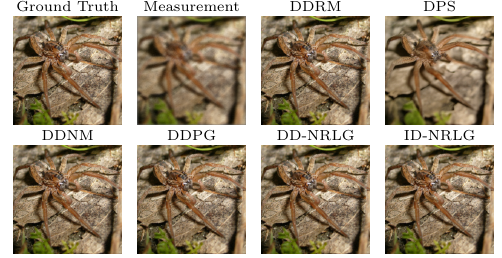} 
    \caption{ ImageNet: Gaussiam Deblurring with noise level 0. }
    \label{figB.10}
\end{figure}

\begin{figure}[t]
    \centering
    \includegraphics[width=0.75\textwidth]{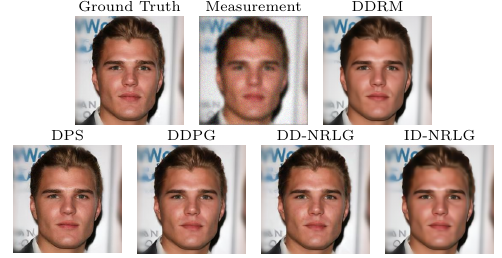} 
    \caption{CelebA-HQ: Gaussian Deblurring with noise level 0.05. }
    \label{figB.11}
\end{figure}

\begin{figure}[t]
    \centering
    \includegraphics[width=0.75\textwidth]{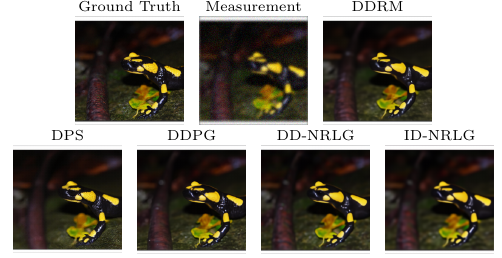} 
    \caption{ ImageNet: Gaussian Deblurring with noise level 0.05. }
    \label{figB.12}
\end{figure}

\begin{figure}[t]
    \centering
    \includegraphics[width=0.75\textwidth]{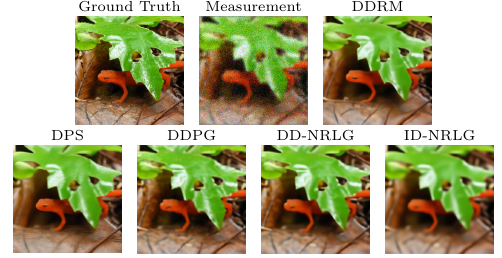} 
    \caption{ CelebA-HQ: Gaussian Deblurring with noise level 0.1. }
    \label{figB.13}
\end{figure}

\begin{figure}[t]
    \centering
    \includegraphics[width=0.75\textwidth]{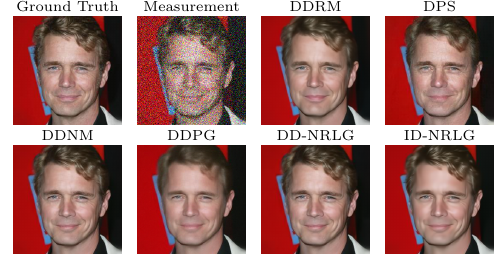} 
    \caption{ ImageNet: Gaussian Deblurring with noise level 0.1. }
    \label{figB.14}
\end{figure}

\begin{figure}[h]
    \centering
    \includegraphics[width=0.75\textwidth]{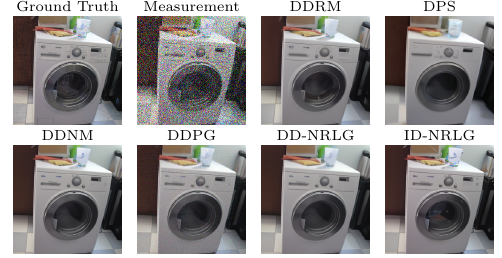} 
    \caption{ImageNet: Denoising with noise level 0.25. }
    \label{figB.15}
\end{figure}

\begin{figure}[h]
    \centering
    \includegraphics[width=0.75\textwidth]{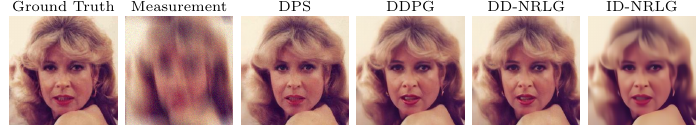} 
    \caption{ CelebA-HQ: Motion Deblurring with noise level 0.05. }
    \label{figB.16}
\end{figure}

\begin{figure}[h]
    \centering
    \includegraphics[width=0.75\textwidth]{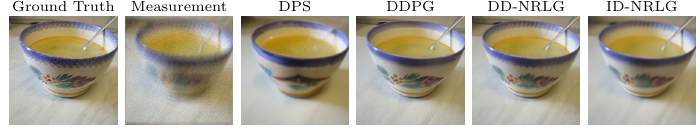}
    \caption{ ImageNet: Motion Deblurring with noise level 0.05. }
    \label{figB.17}
\end{figure}

\begin{figure}[h]
    \centering
    \includegraphics[width=0.75\textwidth]{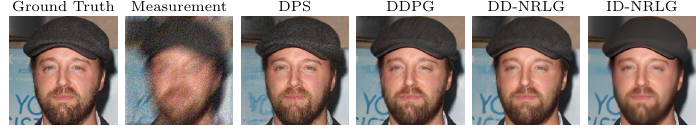} 
    \caption{ CelebA-HQ: Motion Deblurring with noise level 0.1. }
    \label{figB.18}
\end{figure}

\end{document}